\documentclass[preprint,12pt]{elsarticle}

\usepackage{graphicx}
\usepackage{amssymb}
\usepackage{amsthm}
\usepackage{mathtools}
\usepackage{lineno}
\usepackage{mathrsfs}
\usepackage{booktabs, caption, makecell}
\usepackage{threeparttable}
\usepackage{algorithm}
\usepackage{algcompatible}
\usepackage{lipsum}
\usepackage{appendix}

\usepackage[colorlinks,linkcolor=blue]{hyperref}
\usepackage[nameinlink,capitalise]{cleveref}
\usepackage{calc}

\DeclareMathOperator*{\argmax}{arg \, max}
\DeclareMathOperator*{\argmin}{arg \, min}

\theoremstyle{remark}
\newtheorem*{remark}{Remark}

\journal{Elsevier}

\begin{document}

\begin{frontmatter}

\title{Bayesian deep learning with hierarchical prior: Predictions from limited and noisy data}
\author[label1]{Xihaier Luo \corref{cor1}}
\ead{xluo1@nd.edu}

\author[label1]{Ahsan Kareem}
\cortext[cor1]{Corresponding author. 156 Fitzpatrick Hall, Notre Dame, IN 46556, USA.}
\ead{kareem@nd.edu}

\address[label1]{NatHaz Modeling Laboratory, University of Notre Dame,  Notre Dame, IN 46556, United States}

\begin{abstract}
Datasets in engineering applications are often limited and contaminated, mainly due to unavoidable measurement noise and signal distortion. Thus, using conventional data-driven approaches to build a reliable discriminative model, and further applying this identified surrogate to uncertainty analysis remains to be very challenging. In this paper, a deep learning (DL) based probabilistic model is presented to provide predictions based on limited and noisy data. To address noise perturbation, the Bayesian learning method that naturally facilitates an automatic updating mechanism is considered to quantify and propagate model uncertainties into predictive quantities. Specifically, hierarchical Bayesian modeling (HBM) is first adopted to describe model uncertainties, which allows the prior assumption to be less subjective, while also makes the proposed surrogate more robust. Next, the Bayesian inference is seamlessly integrated into the DL framework, which in turn supports probabilistic programming by yielding a probability distribution of the quantities of interest rather than their point estimates. Variational inference (VI) is implemented for the posterior distribution analysis where the intractable marginalization of the likelihood function over parameter space is framed in an optimization format, and stochastic gradient descent method is applied to solve this optimization problem. Finally, Monte Carlo simulation is used to obtain an unbiased estimator in the predictive phase of Bayesian inference, where the proposed Bayesian deep learning (BDL) scheme is able to offer confidence bounds for the output estimation by analyzing propagated uncertainties. The effectiveness of Bayesian shrinkage is demonstrated in improving predictive performance using contaminated data, and various examples are provided to illustrate concepts, methodologies, and algorithms of this proposed BDL modeling technique.
\end{abstract}

\begin{keyword}
Probabilistic modeling \sep Bayesian inference \sep Deep learning \sep Monte Carlo variational inference \sep Bayesian hierarchical modeling \sep Noisy data
\end{keyword}

\end{frontmatter}

\section{Introduction}
\label{sec1}
Applications of data-driven approaches for learning the performance of engineering systems using limited experimental data is hindered by at least three factors. First, the original input-output patterns are often governed by a series of highly nonlinear and implicit partial differential equations (PDEs), and hence approximation of their functional relationships may be proportionally computation demanding \cite{brunton2016discovering, raissi2018hidden}. Secondly, interpolation and extrapolation techniques are usually needed to extract knowledge from acquired data in consideration of only a limited number of sensors used in practice along with the fact that sensor malfunction often occurs in real time. Nevertheless, it is very difficult to establish an accurate discriminative model merely from data, especially when a relatively small dataset is available \cite{soize2010identification, yang2013output}. Thirdly, experimental data is inevitably contaminated by noise from different sources, for example, signal perturbation induced noisy sensing during monitoring. The performance of conventional discriminative algorithms may be noticeably impaired if proper noise reduction has not been performed \cite{javh2018high, natarajan2013learning}. In this context, we present a machine learning based predictive model that is capable of providing high-quality predictions from limited and noisy data.

To date, most machine learning models are deterministic, which implies that a certain input sample $\boldsymbol{x}_i$ is strictly bounded to a point estimator $\hat{\boldsymbol{y}}_i$ notwithstanding the existence of model uncertainties \cite{ghahramani2015probabilistic, kendall2017uncertainties}. Probabilistic modeling, on the other hand, emerges as an attractive alternative on account of its ability to quantify the uncertainty in model predictions, which can prevent a poorly trained model from being overconfident in predictions, and hence helps stakeholders make a more reliable decision \cite{ghahramani2015probabilistic, robert2014machine}. In literature, Gaussian processes (GPs) and generalized polynomial chaos (gPC) are two representative members from the probabilistic modeling family \cite{robert2014machine, rasmussen2004gaussian, ghanem1991stochastic, xiu2003modeling}. From a mathematical standpoint, GPs put a joint Gaussian distribution over input random variables by defining a mean function $E [ \boldsymbol{x} ]$ and a covariance function $Cov [\boldsymbol{x}_i, \boldsymbol{x}_j]$ \cite{rasmussen2004gaussian}. Then, GPs compute hyperparameters of the spatial covariance function and propagte the inherent randomness of $\boldsymbol{x}$ in virtue of the Bayes' theorem. Next, gPC is an effective way to propagate uncertain quantities by means of utilizing a set of random coefficients $\{ \beta_1, \beta_2, \dots, \beta_n \}$ and orthogonal polynomials $\{ \phi_1, \phi_2, \dots, \phi_n \}$. Approximation approaches (e.g. Galerkin projection) are usually used to determine the unknown coefficients $\boldsymbol{\beta}$ and polynomial basis $\boldsymbol{\phi \left( \boldsymbol{x} \right)}$ \cite{xiu2003modeling}. Even though GPs and gPC are capable of computing empirical confidence intervals, the inference complexity may become overwhelming when the number of observations increases (e.g. cubic scaling relationship $\mathcal{O} \left( N^3 \right)$ between the computational complexity and data $\{ \boldsymbol{x}_{i}, \boldsymbol{y}_{i} \}_{i=1}^{N}$ is found in the GPs case \cite{snoek2015scalable}). Furthermore, scaling a GPs model or identifying random coefficients $\boldsymbol{\beta}$ of a gPC model for problems with high-dimensional data remains challenging \cite{soize2010identification, robert2014machine, rasmussen2004gaussian}. 

\begin{figure}[ht]
\centering
\includegraphics[width=0.9\textwidth]{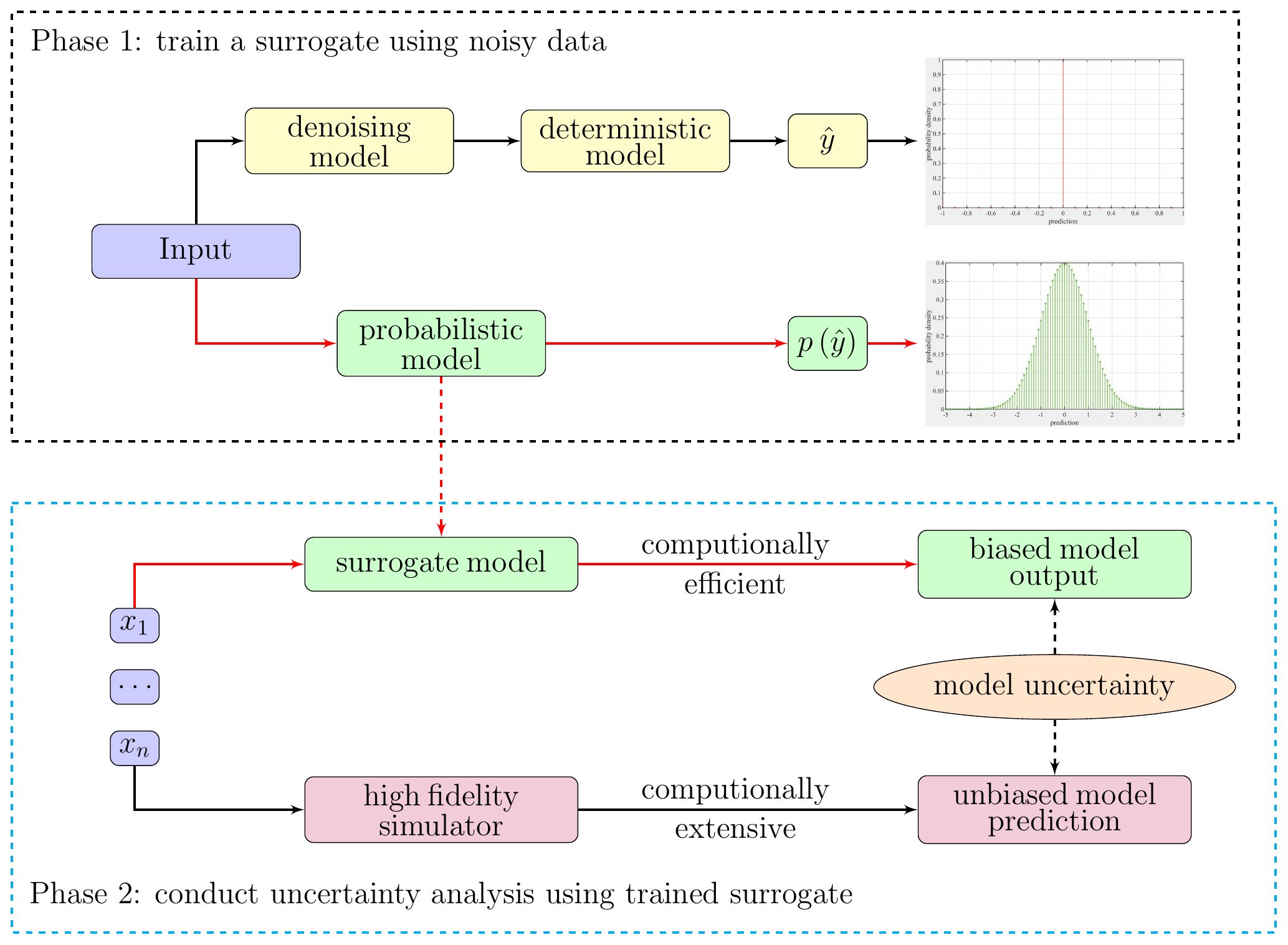}
\caption{The proposed BDL model $\mathcal{M}^{'} \left( \cdot \right)$ aims at reducing the computational burden imposed by the repeated evaluations of the original high-fidelity model $\mathcal{M} \left( \cdot \right)$ in uncertainty analysis. Instead of a point estimate, the BDL model quantifies the model uncertainties by presenting a distribution of values.}
\label{fig: f1}
\end{figure}

On the contrary, deep learning (DL) has differentiated itself within the realm of machine learning for its superior performance in handling large-scale complex systems. With the strong support of high-performance computing (HPC), DL has made significant accomplishments in a wide range of applications such as image recognition, data compression, computer vision, and language processing \cite{goodfellow2016deep, lecun2015deep}. Nonetheless, the same level of application has not been observed for its probabilistic version \cite{mackay1992practical, mackay1995probable, neal2012bayesian}. This paper attempts to bridge the modeling gap between DL and Bayesian learning by presenting a new paradigm Bayesian deep learning (BDL) model. With the aim of developing a surrogate model that can be used to accelerate the uncertainty analysis of engineering systems using noisy data, we focus on three main aspects in probabilistic modeling (See \cref{fig: f1}).

First, Bayesian statistical inference encodes the subjective beliefs, whereas prior distributions are imposed on model parameters to represent the initial uncertainty. A conventional DL supported surrogate $\mathcal{M}^{'} \left( \cdot \right)$, however, can be captivating and confusing in equal measure at the same time. Owning a deeply structured network architecture allows $\mathcal{M}^{'} \left( \cdot \right)$ to approximate a wide variety of functions, but also makes model parameters hard to interpret, which in turn increases the difficulty of choosing a reasonable prior distribution for $\mathcal{M}^{'} \left( \cdot \right)$ \cite{lee2004bayesian, nalisnick2018priors}. In \cite{lee2004bayesian}, Lee shows noninformative prior (e.g. Jeffreys prior) that bears objective Bayes properties is liable to be misled by the variability in data. And using the Fisher information matrix to compute a Jeffreys prior for a large network architecture can be computationally prohibitive \cite{jeffreys1946invariant}. On the informative prior side, zero-centered Gaussian is, not surprisingly, extensively explored in early work on Bayesian neural networks (BDL) on account of its flexibility in implementation as well as its natural regularization mechanism \cite{mackay1992practical, mackay1995probable}, where the quadratic penalty for BDL parameters alleviates the overfitting problem. Later in \cite{neal2012bayesian}, Neal points out that employing a heavy-tailed distribution (e.g. Cauchy distribution) to represent prior knowledge can provide a more robust shrinkage estimator and diminish the effects of outlying data. Nonetheless, Cauchy distribution is difficult to implement because it does not have finite moments \cite{nalisnick2018priors}. To develop a prior model that is more amenable to reform and work with, we investigate the efficacy of applying hierarchical Bayesian modeling (HBM) to define the prior distribution. And it is found that HBM can efficiently ameliorate the prior assumptions induced model performance variance by diffusing the influences that are cascaded down from the top level, hence allowing a relatively more robust distribution model.

Secondly, the determination of the posterior distribution $p \left( \boldsymbol{\omega} | \boldsymbol{\mathcal{D}} \right)$ requires integrating model parameters $\{ \omega_{1}, \omega_{2}, \dots, \omega_{n} \}$ out of the likelihood function. Unfortunately, performing numerical integration in BDL's parameter space is always computationally intractable as a neural networks model is commonly configured with hundreds/thousands of parameters \cite{gal2016uncertainty, li2018approximate}. Approximation methods that can be broadly classified into the sampling method and the optimization method have been introduced to alleviate such computational bottleneck \cite{zhu2017big}. For the the sampling method, Markov Chain Monte Carlo (MCMC) has been explored in early work to calculate the posterior probabilities \cite{mackay1995probable, nalisnick2018priors, blundell2015weight}. The main idea of MCMC is to produce numerical samples from the posterior distribution by simulating a discrete but dependent Markov chain $\{ \boldsymbol{\omega}_{i} \}_{i=1}^{M}$ on the state space, where $\pi \left( \boldsymbol{\omega} \right) \approx p \left( \boldsymbol{\omega} | \boldsymbol{\mathcal{D}} \right)$. The sufficient condition that ensures the stationary distribution converges to the target posterior distribution requires the transition kernel $T \left( \cdot \right)$ to have detailed balance properties $p \left( \boldsymbol{\omega} | \boldsymbol{\mathcal{D}} \right) T \left( \boldsymbol{\omega}^{'} \rightarrow \boldsymbol{\omega} \right) = p \left( \boldsymbol{\omega}^{'} | \boldsymbol{\mathcal{D}} \right) T \left( \boldsymbol{\omega} \rightarrow \boldsymbol{\omega}^{'} \right)$. However, the convergence of MCMC algorithms can be extremely slow in the presence of large datasets since the burn-in process to eliminate the initialization bias is greatly extended \cite{robert2014machine, li2018approximate}. More recently, variational inference (VI) method has been employed for inferring an intractable posterior distribution where the probabilistic inference problem is cast in a deterministic optimization form \cite{blei2017variational, jordan1999introduction, hoffman2013stochastic}. A proxy probability distribution $q \left( \boldsymbol{\omega} \right)$, which is formally explicit and computationally efficient, is introduced to approximate the true posterior distribution $p \left( \boldsymbol{\omega} | \boldsymbol{\mathcal{D}} \right)$. Compared to the sampling method, VI has the advantage of approximating non-conjugate distributions by virtue of optimizing an explicit objective function \cite{li2018approximate, zhu2017big}, and VI solves the intractable integrals in a more efficient manner on account of off-the-peg optimization algorithms can be seamlessly adapted to the minimization problem \cite{blei2017variational, hoffman2013stochastic, ranganath2014black}. Unfortunately, the differentiation of the objective function regarding the proxy posterior involves the determination of the expectations with respect to the variational parameters where Monte Carlo gradient estimator may give a high variance \cite{kingma2013auto, rezende2014stochastic}. To address this issue, we reparameterize our objective function by introducing a set of auxiliary variables. It should be noted that such reparameterization would not only yield an unbiased estimator of the objective function but also provides an efficient approximation of variational parameters via permitting the use of stochastic gradient descent (SGD) in optimization. 

Lastly, Monte Carlo (MC) method is used in the predictive phase of Bayesian inference. MC method draws numerical samples from the proxy probability distribution $q \left( \boldsymbol{\omega} \right)$, builds a predictive probability distribution of new data, and assigns a confidence level to the model prediction for representing model uncertainty. The following outline of this paper is intended as: \cref{sec2} gives a brief introduction of surrogate modeling using deep neural networks. \cref{sec3} describes the proposed BDL model in detail. In \cref{sec4}, various examples are provided to demonstrate the effectiveness of BDL in dealing with noisy data. Finally, \cref{sec5} draws major conclusions.  

\section{Deterministic modeling: a deep learning framework}
\label{sec2}
\subsection{Neural networks based surrogate model}
\label{sec21}
In the context of supervised learning \cite{robert2014machine, lecun2015deep}, let $\boldsymbol{x} = [x_1, x_2, \dots, x_m] \in \mathbb{R}^m$ denote an input vector, and the corresponding output vector $\boldsymbol{y} = [y_1, y_2, \dots, y_n] \in \mathbb{R}^n$ is estimated by a computationally intensive model $\mathcal{M} \left( \cdot \right)$ (e.g. a large finite element model). We are interested in using neural networks to approximate functional relationships $\mathcal{F} \left( \cdot \right)$ between $\boldsymbol{x}$ and $\boldsymbol{y}$.

\begin{equation}
\label{eq: sec2_1}
\boldsymbol{y} = \mathcal{F} \left( \boldsymbol{x} \right) \xrightarrow{\approx} \hat{\boldsymbol{y}} = \hat{\mathcal{F}} \left( \boldsymbol{x} \right)
\end{equation}

where $\hat{\mathcal{F}} \left( \cdot \right)$ is the mathematical expression of neural networks based surrogate $\mathcal{M}^{'} \left( \cdot \right)$, and theoretically it can be proportionately broken down as:

\begin{equation}
\label{eq: sec2_2}
\hat{\mathcal{F}} \left( \boldsymbol{x} \right) = \hat{f}^{K} \circ \hat{f}^{K-1} \circ \dots  \hat{f}^{1} \left( \boldsymbol{x} \right)
\end{equation}

with $K$ denoting the layer number, and $\circ$ symbolizing the functional composition operation which is defined as \cite{nasrabadi2007pattern}:

\begin{equation}
\label{eq: sec2_3}
\hat{f}^{i} \circ \hat{f}^{j} \doteq \hat{f}^{i} \left( \hat{f}^{j} \left( \cdot \right) \right)
\end{equation}

Each function in the sequence $\hat{f}^{i} \left( \cdot \right), i=1, 2, \dots, K$ contains two steps, where the first step is identical to a linear regression:

\begin{equation}
\label{eq: sec2_4}
\boldsymbol{z}^{i} = \hat{f}^{i}_{1} \left( \boldsymbol{x}^{i} \right) = \boldsymbol{\omega}^{i} \boldsymbol{x}^{i} + \boldsymbol{b}^{i} 
\end{equation}

In \cref{eq: sec2_4}, $\boldsymbol{x}^{i}$ is the input vector of the $i^{th}$ layer, $\boldsymbol{\omega}^{i}$ is the weight matrix, and $\boldsymbol{b}^{i}$ is the $i^{th}$ bias term. For the sake of brevity, $\boldsymbol{b}^{i}$ can be integrated into $\boldsymbol{\omega}^{i}$ by introducing an additional input variable $x_0^{i} = 1$. For the rest of the paper, let $\boldsymbol{\omega}$ be a tensor containing all model parameters \cite{nasrabadi2007pattern, abadi2016tensorflow}. Next, $\hat{f}^{i} \left( \cdot \right)$ applies an element-wise nonlinear transformation to the intermediate output $\boldsymbol{z}^{i}$ in the second step:

\begin{equation}
\label{eq: sec2_6}
\hat{\boldsymbol{y}}^{i} = \hat{f}^{i}_{2} \left( \boldsymbol{x}^{i} \right) = \sigma \left( \boldsymbol{z}^{i} \right)
\end{equation}

where $\sigma \left( \cdot \right)$ is often referred to as the activation function \cite{goodfellow2016deep, lecun2015deep}. Selection of $\sigma \left( \cdot \right)$ directly depends on the characteristics of $\mathcal{M} \left( \cdot \right)$. Moreover, a network architecture is deemed to be deep when $K > 3$ \cite{goodfellow2016deep}. Thus, a deep learning framework can be effectively built by increasing the composition size $K$.

\subsection{Probabilistic interpretation of $L^2$ loss function}
\label{sec22}
Consider a parameterised deep neural network model $\hat{\boldsymbol{y}} = \hat{\mathcal{F}}_{\boldsymbol{\omega}} \left( \boldsymbol{x} \right)$ described in \cref{sec21} and a training dataset $\boldsymbol{\mathcal{D}} = \{ \boldsymbol{x}_{i}, \boldsymbol{y}_{i} \}_{i=1}^{N}$, the next step is to find an optimal $\boldsymbol{\omega}^{\star}$ such that the surrogate $\hat{\mathcal{F}}_{\boldsymbol{\omega}} \left( \boldsymbol{x} \right)$ best describes the data $\boldsymbol{\mathcal{D}}$. In this regard, a loss function $\mathcal{L} \left( \cdot \right)$ that measures the error between the predicted value $\hat{\boldsymbol{y}}$ and the expected result $\boldsymbol{y}$ is defined. Then, $\hat{\mathcal{F}}_{\boldsymbol{\omega}} \left( \boldsymbol{x} \right)$ is trained to approximate $\mathcal{F} \left( \boldsymbol{x} \right)$ by minimizing the empirical loss through tuning model parameters $\boldsymbol{\omega}$:

\begin{equation}
\label{eq: sec2_7}
\boldsymbol{\omega}^{\star} = \argmin \sum_{i=1}^{n} \mathcal{L} \left( \boldsymbol{y}_i, \hat{\mathcal{F}} \left( \boldsymbol{x}_i \right) \right)
\end{equation}

From a probabilistic modeling perspective, the interest of loss function $\mathcal{L} \left( \cdot \right)$ is in the probability distribution of $\boldsymbol{y}$ as a funciton of $\boldsymbol{x}$:

\begin{equation}
\label{eq: sec2_8}
\mathcal{L} \left( \boldsymbol{y}_i, \hat{\mathcal{F}} \left( \boldsymbol{x}_i \right) \right) = p \left( \boldsymbol{\mathcal{D}} | \boldsymbol{\omega} \right) = p \left( \boldsymbol{y} | \boldsymbol{x}, \boldsymbol{\omega} \right)
\end{equation}

where the model parameters can be learned by the method of maximum likelihood estimate (MLE) \cite{robert2014machine, gal2016uncertainty}, which searches an estimator for $\boldsymbol{\omega}$ that maximizes the likelihood term $\boldsymbol{\omega}^{MLE} = \argmax p \left( \boldsymbol{\mathcal{D}} | \boldsymbol{\omega} \right)$. Let noise term $\boldsymbol{\epsilon}$ be independent and identically distributed (i.i.d.), the probability density associated with paired observations under Gaussian assumption can be expressed as:

\begin{equation}
\label{eq: sec2_9}
\begin{aligned}
\mathcal{L} \left( \boldsymbol{y}_i, \hat{\mathcal{F}} \left( \boldsymbol{x}_i \right) \right) &  = p \left( \boldsymbol{\mathcal{D}} | \boldsymbol{\omega}, \tau_{\boldsymbol{\epsilon}}  \right) = \prod_{i=1}^{n} \mathcal{N} \left( \boldsymbol{y}_i \, | \, \hat{\mathcal{F}} \left( \boldsymbol{x}_i \right),  \tau_{\boldsymbol{\epsilon}}^{-1} \right) \\
& = \left( \frac{1}{\left( 2 \pi \boldsymbol{\sigma}_{\boldsymbol{\epsilon}}^2 \right)^n}\right)^{\frac{1}{2}} \prod_{i=1}^{n} \exp \left( - \frac{1}{2 \boldsymbol{\sigma}_{\boldsymbol{\epsilon}}^2} \left( \boldsymbol{y}_i - \hat{\mathcal{F}} \left( \boldsymbol{x}_i \right)  \right)^2 \right)
\end{aligned}
\end{equation}

Usually, the numerical implementation of the MLE method performs the minimization problem of \cref{eq: sec2_7} in a logarithmic scale:

\begin{equation}
\label{eq: sec2_10}
\boldsymbol{\omega}^{\star} = \argmin \sum_{i=1}^{n} \left( - \frac{1}{2 \boldsymbol{\sigma}_{\boldsymbol{\epsilon}}^2} \left( \boldsymbol{y}_i - \hat{\mathcal{F}} \left( \boldsymbol{x}_i \right)  \right)^2 \right) - \frac{n}{2} \log \left( 2 \pi \boldsymbol{\sigma}_{\boldsymbol{\epsilon}}^2 \right)
\end{equation}

where the precision term $\tau_{\boldsymbol{\epsilon}}$ is determined by minimizing the negative log likelihood:

\begin{equation}
\label{eq: sec2_11}
\tau_{\boldsymbol{\epsilon}}^{MLE} = \left( \frac{1}{n N_{\boldsymbol{\omega}}} \sum_{i=1}^{n} \left( \boldsymbol{y}_i - \hat{\mathcal{F}} \left( \boldsymbol{x}_i \right) \right)^2 \right)^{-1}
\end{equation}

Furthermore, \cref{eq: sec2_9} can be further simplified under homoscedastic conditions:

\begin{equation}
\label{eq: sec2_12}
\boldsymbol{\omega}^{\star} = \argmin \sum_{i=1}^{n} \left( \boldsymbol{y}_i - \hat{\mathcal{F}} \left( \boldsymbol{x}_i \right) \right)^2
\end{equation}

Hence, the probability density based loss function coincides with the well-known mean squared error (MSE).

\subsection{Stochastic optimization for updating model parameters}
\label{sec23}
Gradient based optimization is one of the most popular algorithms to optimize neural networks:

\begin{equation}
\label{eq: sec2_13}
\boldsymbol{\omega}_{t+1} \gets \boldsymbol{\omega}_{t} + \eta \, \nabla \mathcal{L} \left( \boldsymbol{\omega} \right)
\end{equation}

where $\eta$ is generally known as the leraning rate that follows the Robbins-Monro conditions. The objective function stated in \cref{eq: sec2_12} indicates $\mathcal{O} \left( N \right)$ operations are required to compute $\mathcal{L} \left( \cdot \right)$ and $\nabla \mathcal{L} \left( \cdot \right)$ respectively, which may be computationally demanding for a large dataset. Therefore, stochastic gradient descent (SGD) is considered \cite{goodfellow2016deep, lecun2015deep, nasrabadi2007pattern}, where a random vector $g \left( \boldsymbol{x} \right)$ is defined to calculate the gradients \cite{robbins1985stochastic}. With a restricted magnitude of stochastic gradients $\mathbb{E} || g \left( \boldsymbol{x} \right) ||^2 \leqslant N^2$ and a bounded variance $\mathbb{E} || g \left( \boldsymbol{x} \right) - \nabla \mathcal{L} \left( \boldsymbol{x} \right) ||^2 \leqslant \sigma^2$ \cite{robbins1985stochastic}, SGD can efficiently update model parameters by constructing a noisy natural gradient:

\begin{equation}
\label{eq: sec2_14}
\nabla \mathcal{L} \left( \cdot \right) = \mathbb{E} [ g \left( \boldsymbol{x} \right) ]
\end{equation}

In particular, adaptive moment estimation (ADAM) \cite{kingma2014adam} that computes adaptive learning rates for $\boldsymbol{\omega}$ is adopted, and the corresponding $g \left( \boldsymbol{x} \right)$ takes the expression of:

\begin{equation}
\label{eq: sec2_15}
g \left( \boldsymbol{x} \right) = \dfrac{\hat{M}_t}{\sqrt{\hat{V}_t} + \epsilon} = \dfrac{M_t}{1 - \beta^t_1} / \left( \sqrt{\dfrac{V_t}{1 - \beta^t_2}} + \epsilon \right)
\end{equation}

where $M_t$ and $V_t$ are estimates of the mean and variance of the gradients respectively. In ADAM, they are updated as follows \cite{kingma2014adam}:

\begin{equation}
\label{eq: sec2_16}
\begin{aligned}
M_t &= \beta_1 M_{t-1} + (1 - \beta_1) \mathcal{L}_{t} \left( \cdot \right) \\  
V_t &= \beta_2 V_{t-1} + (1 - \beta_2) \mathcal{L}_{t} \left( \cdot \right)^2  
\end{aligned}
\end{equation}

It should be noted that the expresssion of \cref{eq: sec2_15} is an unbiased estimation of the exact gradient, and its calculation only depends on one data point.

\section{Probabilistic modeling: a Bayesian approach}
\label{sec3}
In this section, the aforementioned deterministic DL surrogate is enhanced to account for model uncertainties by the integration of Bayesian inference. Overall, Bayesian learning includes three steps: (1) establish prior beliefs about uncertain parameters; (2) compute the posterior distribution via Bayes' rule; and (3) use the predictive distribution to determine a yet unobserved data point. 

\subsection{Prior representation: Bayesian hierarchical modelling}
\label{sec31}
To begin with, let $\mathcal{U}_{\mathcal{E}}$ and $\mathcal{U}_{\mathcal{A}}$ represent the epistemic uncertainty and aleatory uncertainty respectively \cite{kendall2017uncertainties}. Prior information of $\mathcal{U}_{\mathcal{E}}$ and $\mathcal{U}_{\mathcal{A}}$ is initially encapsulated in a probability distribution function form. For the epistemic uncertainty, prior distributions are imposed on model parameters $\boldsymbol{\omega}$ \cite{mackay1992practical, mackay1995probable, neal2012bayesian}:

\begin{equation}
\label{eq: sec3_1}
\boldsymbol{\omega} \sim p \left( \boldsymbol{\omega} \right)
\end{equation}

Practical applications imply that the prior distribution $p \left( \boldsymbol{\omega} \right)$ should not be too restrictive on account of the limited prior information about $\boldsymbol{\omega}$ \cite{nalisnick2018priors}. For this reason, hierarchical Bayesian modeling (HBM) method, which introduces a vector of hyperparameters $\boldsymbol{\eta} = [\eta_1, \eta_2, \dots, \eta_n]$ to the prior distribution, is employed to reduce subjective information induced undue influence on $p \left( \boldsymbol{\omega} \right)$ \cite{robert2014machine}. Consequently, the marginal prior can be obtained by integrating out $\boldsymbol{\eta}$ through the sum rule:

\begin{equation}
\label{eq: sec3_2}
p \left( \boldsymbol{\omega} \right) = \int p \left( \boldsymbol{\omega}, \boldsymbol{\eta} \right) d \boldsymbol{\eta}
\end{equation}

where the joint probability distribution can be further expressed as a product of a set of conditional distributions via applying the product rule:

\begin{equation}
\label{eq: sec3_3}
p \left( \boldsymbol{\omega}, \boldsymbol{\eta} \right) = p \left( \boldsymbol{\omega} | \boldsymbol{\eta} \right) p \left( \boldsymbol{\eta} \right)
\end{equation}

For probabilistic modeling, model parameters in each layer of a BDL model are often assumed to follow a factorized multivariate Gaussian distribution:

\begin{equation}
\label{eq: sec3_4}
p \left( \boldsymbol{\omega} \right) = \prod_{i=1}^{K} p \left( \boldsymbol{\omega}_{i} \, | \, \boldsymbol{\mu}_{\boldsymbol{\omega}_{i}}, \tau_{\boldsymbol{\omega}_{i}} \right) = \prod_{i=1}^{K} \mathcal{N} \left( \boldsymbol{\mu}_{\boldsymbol{\omega}_{i}}, \tau_{\boldsymbol{\omega}_{i}}^{-1} \, \boldsymbol{I} \right)
\end{equation}

At the first hierarchy stage, let $\boldsymbol{\mu}_{\boldsymbol{\omega}} = \boldsymbol{0}$ and $\tau_{\boldsymbol{\omega}}$ be a Gamma random variable for instance. Hence, HBM breaks the prior distribution down to:

\begin{equation}
\label{eq: sec3_5}
p \left( \boldsymbol{\omega} \, | \, \tau_{\boldsymbol{\omega}}, \alpha_{\tau}, \beta_{\tau} \right) = p \left( \boldsymbol{\omega} \, | \, \tau_{\boldsymbol{\omega}} \right) p \left( \tau_{\boldsymbol{\omega}} | \alpha_{\tau}, \beta_{\tau} \right) \, \, \, \text{where} \, \, \, \alpha_{\tau} > 0, \beta_{\tau} > 0
\end{equation}

with $\alpha_{\tau}$ and $\beta_{\tau}$ denoting the shape parameter and rate parameter of the Gamma distribution respectively. Using \cref{eq: sec3_2} and  \cref{eq: sec3_3}, the prior distribution can be reformulated as:

\begin{equation}
\label{eq: sec3_6}
\begin{aligned}
p \left( \boldsymbol{\omega} \, | \, \alpha_{\tau}, \beta_{\tau} \right)  & = \int_{0}^{\infty} p \left( \boldsymbol{\omega} \, | \, \tau_{\boldsymbol{\omega}}, \alpha_{\tau}, \beta_{\tau} \right)  d \tau_{\boldsymbol{\omega}} = \mathcal{S}t \left( \boldsymbol{0}, \frac{\alpha_{\tau}} {\beta_{\tau}}, 2 \alpha_{\tau} \right)
\end{aligned}
\end{equation}

In \cref{eq: sec3_6}, $\mathcal{S}t \left( \cdot \right)$ characterizes the student's t-distribution which is capable of providing heavier tails than Gaussian distribution.

\begin{remark}[1]
HBM grants a more impartial prior distribution by allowing the data to speak for itself \cite{robert2014machine}, and it admits a more general modeling framework where the hierarchical prior becomes direct prior when the hyperparameters are modeled by a Dirac delta function (e.g. using $\delta \left( x - \tau_{\boldsymbol{\omega}} \right) $ to describe the precision term in \cref{eq: sec3_4}). In addition, HBM offers the flexibility to work with a wide range of probability distributions, and even directly provides an analytical solution for some of the most popular choices such as Laplace, Gaussian, and student's t-distribution \cite{nalisnick2018priors}.
\end{remark}

On the other hand, homoscedastic noise $\boldsymbol{\epsilon}$ that is independent of the input data $\mathbb{V}ar [ \boldsymbol{\epsilon} | \left( \boldsymbol{x}_i, \boldsymbol{y}_i \right) ]  = \boldsymbol{\sigma}_{\boldsymbol{\epsilon}}^2 \, \forall \, \left( \boldsymbol{x}_i, \boldsymbol{y}_i \right) \in \boldsymbol{\mathcal{D}}$ is added to the output in consideration of the aleatory uncertainty $\mathcal{U}_{\mathcal{A}}$ which cannot be explained away by accepting more samples $\{ \boldsymbol{x}, \boldsymbol{y} \}$. Besides, additive noise term $\boldsymbol{\epsilon}$ guarantees a tractable likelihood for the probabilistic model $\mathcal{M}^{'} \left( \cdot \right)$ where $\boldsymbol{\epsilon}$ is most commonly modeled as a Gaussian process:

\begin{equation}
\label{eq: sec3_7}
p \left( \boldsymbol{\epsilon} \, | \, \boldsymbol{\mu}_{\boldsymbol{\epsilon}}, \boldsymbol{\sigma}_{\boldsymbol{\epsilon}}^2 \right) = \mathcal{N} \left( \boldsymbol{\mu}_{\boldsymbol{\epsilon}}, \boldsymbol{\sigma}_{\boldsymbol{\epsilon}}^2 \right)
\end{equation}

Let $\boldsymbol{\mu}_{\boldsymbol{\epsilon}} = \boldsymbol{0}$ and $\boldsymbol{\sigma}_{\boldsymbol{\epsilon}}$ be a constant, the output vector hereby follows:

\begin{equation}
\label{eq: sec3_8}
\boldsymbol{y} \sim \mathcal{N} \left( \boldsymbol{y} \, | \, \mathbb{E}_{p \left( \boldsymbol{\omega} \, | \, \boldsymbol{\mathcal{D}} \right)} [ \hat{\mathcal{F}} \left( \boldsymbol{x} \right) ], \tau_{\boldsymbol{\epsilon}} \boldsymbol{I} \right)
\end{equation}

\begin{figure}[ht]
\centering
\includegraphics[width=0.7\textwidth]{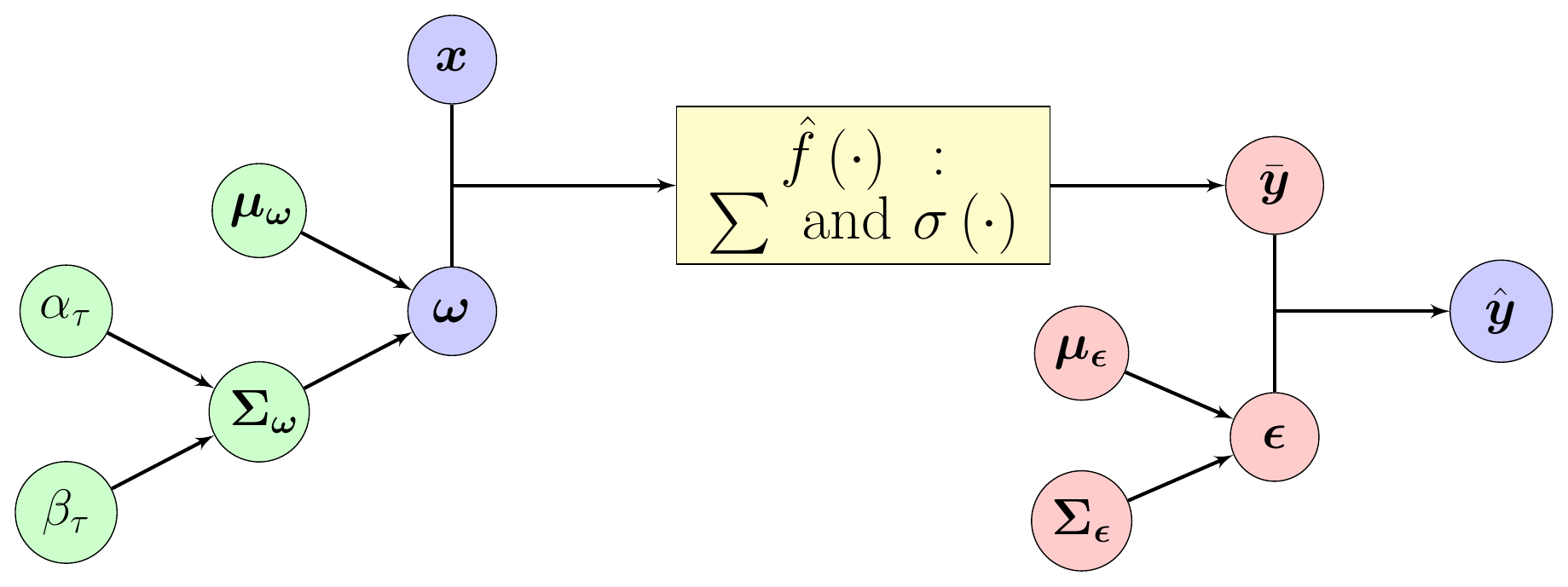}
\caption{The architecture of Bayesian deep learning with hierarchical prior.}
\label{fig: f2}
\end{figure}

A graphical model representation for the aforestated hierarchical prior as well as an illustration of the BDL model is given in \cref{fig: f2}.

\subsection{Posterior approximation: variational inference}
\label{sec32}
After defining a hierarchical prior distribution for the proposed probabilistic model $\mathcal{M}^{'} \left( \cdot \right)$, the next step is to infer the posterior distribution, which reflects the updated parameter information. In the Bayesian formalism, the joint posterior distribution $p \left( \boldsymbol{\omega}, \boldsymbol{\eta} \, | \, \boldsymbol{\mathcal{D}} \right)$ is calculated by:

\begin{equation}
\label{eq: sec3_9}
p \left( \boldsymbol{\omega}, \boldsymbol{\eta} \, | \, \boldsymbol{\mathcal{D}} \right) = \frac{p \left( \boldsymbol{\mathcal{D}} \, | \, \boldsymbol{\omega}, \boldsymbol{\eta} \right) p \left( \boldsymbol{\omega} \, | \, \boldsymbol{\eta} \right) p \left( \boldsymbol{\eta} \right)}{p \left( \boldsymbol{\mathcal{D}} \right)}
\end{equation}

The marginal posterior distribution $p \left( \boldsymbol{\omega} \, | \, \boldsymbol{\mathcal{D}} \right)$ can be further determined by integrating out the joint posterior distribution, and the denominator of \cref{eq: sec3_9} is often referred to as the model evidence that takes the form of: 

\begin{equation}
\label{eq: sec3_10}
p \left( \boldsymbol{\mathcal{D}} \right) = \int_{\boldsymbol{\omega}} p \left( \boldsymbol{\mathcal{D}} \, | \, \boldsymbol{\omega} \right) p \left( \boldsymbol{\omega} \right) d \boldsymbol{\omega}
\end{equation}

In most cases, estimation of \cref{eq: sec3_10} it is computationally intractable as numerical integration requires a considerable number of samples if the parameter space $\mathcal{W}$ is very high \cite{neal2012bayesian}. To overcome this integration problem, variational inference (VI) is adopted so that Bayesian inference can proceed efficiently \cite{blei2017variational, hoffman2013stochastic}.

\begin{remark}[2]
Different from the method of maximum a posteriori (MAP) which captures the mode of a posterior distribution \cite{neal2012bayesian}, the objective for the posterior approximation at this place is to find a computationally efficient replacement of the true posterior distribution, so that numerical samples can be easily accessible in the predictive analysis.
\end{remark}

\subsubsection{Objective function: evidence lower bound}
\label{sec321}
In VI, a family of proxy distributions parameterized by $\boldsymbol{\xi}$ is posited to approximate the true posterior distribution:

\begin{equation}
\label{eq: sec3_11}
p \left( \boldsymbol{\omega} \, | \, \boldsymbol{\mathcal{D}} \right) \approx q_{\boldsymbol{\xi}} \left( \boldsymbol{\omega} \right) \in \boldsymbol{\Xi} 
\end{equation}

VI attempts to make $q_{\boldsymbol{\xi}} \left( \boldsymbol{\omega} \right)$ looks as close as possible to $p \left( \boldsymbol{\omega} | \boldsymbol{\mathcal{D}} \right)$ via refining $\boldsymbol{\xi}$, and one typical interpretation of the \emph{closeness} between two probability distributions is the Kullback-Leibler (KL) divergence \cite{jordan1999introduction, hoffman2013stochastic}. Therefore, VI casts the approximation problem in an optimization form, where the objective function can be expressed as: 

\begin{equation}
\label{eq: sec3_12}
\begin{aligned}
p \left( \boldsymbol{\omega} \, | \, \boldsymbol{\mathcal{D}} \right) \simeq q_{\boldsymbol{\xi}^{\star}} \left( \boldsymbol{\omega} \right) & = \argmin \mathbb{K} \mathbb{L} \left( q_{\boldsymbol{\xi}} \left( \boldsymbol{\omega} \right) | p \left( \boldsymbol{\omega} | \boldsymbol{\mathcal{D}} \right) \right) \\
& = \argmin \int q_{\boldsymbol{\xi}} \left( \boldsymbol{\omega} \right) \log \frac{q_{\boldsymbol{\xi}} \left( \boldsymbol{\omega} \right)}{p \left( \boldsymbol{\omega} | \boldsymbol{\mathcal{D}} \right)} d \boldsymbol{\omega}
\end{aligned}
\end{equation}

Instead of minimizing the KL divergence, we can equivalently maximize the evidence lower bound (ELBO) $\mathcal{L} \left( \boldsymbol{\omega} \right)$ \cite{hoffman2013stochastic}: 

\begin{equation}
\label{eq: sec3_13}
\begin{aligned}
q_{\boldsymbol{\xi}^{\star}} \left( \boldsymbol{\omega} \right) & = \mathcal{L} \left( \boldsymbol{\omega} \right) = \argmax  = \argmax \mathbb{E}_{q_{\boldsymbol{\xi}} \left( \boldsymbol{\omega} \right)} \left[ \log p \left( \boldsymbol{\omega} , \boldsymbol{\mathcal{D}} \right) - \log q_{\boldsymbol{\xi}} \left( \boldsymbol{\omega} \right) \right] \\
& = \argmax \mathbb{E}_{q_{\boldsymbol{\xi}} \left( \boldsymbol{\omega} \right)} \left[  \log p \left( \boldsymbol{\mathcal{D}} \, | \, \boldsymbol{\omega} \right) \right] - \mathbb{K} \mathbb{L} \left( q_{\boldsymbol{\xi}} \left( \boldsymbol{\omega} \right) \, | \, p \left( \boldsymbol{\omega} \right) \right)
\end{aligned}
\end{equation}

The first conditional log-likelihood term in \cref{eq: sec3_13} is usually referred as to the data term \cite{hoffman2013stochastic, ranganath2014black}. It compels the posterior distribution to explain data $\boldsymbol{\mathcal{D}}$ by maximizing the expected log-likelihood. Mini-batch optimization method is implemented to efficiently offer an unbiased stochastic estimator of the log-likelihood:

\begin{equation}
\label{eq: sec3_14}
\log p \left( \boldsymbol{\mathcal{D}} | \boldsymbol{\omega} \right) = \sum_{i=1}^{N} \log p \left( \boldsymbol{y}_{i} | \boldsymbol{x}_{i}, \boldsymbol{\omega} \right) \approx \frac{N}{M}\sum_{i=1}^M \log p \left( \boldsymbol{y}_{i} | \boldsymbol{x}_{i}, \boldsymbol{\omega} \right)
\end{equation}

where $M$ is a subset of $N$. Noticeably, besides accelerating the computational process, mini-batch optimization owns a higher model updating frequency that allows for a more robust convergence, and hence increases the chance of avoiding local minimum \cite{robbins1985stochastic, kingma2014adam}. Meanwhile, mean field variational inference (MFVI) method \cite{blei2017variational, hoffman2013stochastic, ranganath2014black} is adopted to control the computational complexity of the second term in \cref{eq: sec3_13}:

\begin{equation}
\label{eq: sec3_15}
q_{\boldsymbol{\xi}} \left( \boldsymbol{\omega} \right) = \prod_{i=1}^{K} q_{\boldsymbol{\xi}_{i}} \left( \boldsymbol{\omega}_{i} \right)
\end{equation}

The variational distribution is represented by a layer-wise factorized distribution where each factor is determined by its own variational parameter:

\begin{equation}
\label{eq: sec3_16}
q_{\boldsymbol{\xi}_{i}} \left( \boldsymbol{\omega}_{i} \right) = \frac{\exp \left( \int \log p \left( \boldsymbol{\mathcal{D}}, \boldsymbol{\omega} \right) \prod_{j \neq i} q_{\boldsymbol{\xi}_{j}} \left( \boldsymbol{\omega}_{j} \right) d \boldsymbol{\omega}_{j} \right)}{\int \exp \left( \int \log p \left( \boldsymbol{\mathcal{D}}, \boldsymbol{\omega} \right) \prod_{j \neq i} q_{\boldsymbol{\xi}_{j}} \left( \boldsymbol{\omega}_{j} \right) d \boldsymbol{\omega}_{j} \right) d \boldsymbol{\omega}_{i}}
\end{equation}

Substituting \cref{eq: sec3_14} and \cref{eq: sec3_15} back to \cref{eq: sec3_13}, the objective function of ELBO can be rewritten into:

\begin{equation}
\label{eq: sec3_17}
\begin{aligned}
\mathcal{L} \left( \boldsymbol{\omega}, \boldsymbol{\xi} \right) & = \int \frac{N}{M}\sum_{i=1}^M \log p \left( \boldsymbol{y}_{i} | \boldsymbol{x}_{i}, \boldsymbol{\omega} \right) \prod_{j =1}^{K} q_{\boldsymbol{\xi}_{j}} \left( \boldsymbol{\omega}_{j} \right) d \boldsymbol{\omega} \\
& - \sum_{j=1}^{K} \int q_{\boldsymbol{\xi}_{j}} \left( \boldsymbol{\omega}_{j} \right) \log q_{\boldsymbol{\xi}_{j}} \left( \boldsymbol{\omega}_{j} \right) d \boldsymbol{\omega}_{j}
\end{aligned}
\end{equation}

where the iteration of variational distribution for model parameters terminates when the convergence criteria is satisfied.

\subsubsection{Gradients computation: stochastic gradient variational Bayes}
\label{sec322}
Among the many techniques developed for solving optimization problems, gradient-based optimization method reliably tackles the EBLO maximization problem stated in \cref{eq: sec3_13} in an efficient manner \cite{rezende2014stochastic}. For the sake of brevity, let:

\begin{equation}
\label{eq: sec3_18}
\mathcal{A} \left( \boldsymbol{\omega}, \boldsymbol{\xi} \right) = \log p \left( \boldsymbol{\omega} , \boldsymbol{\mathcal{D}} \right) - \log q_{\boldsymbol{\xi}} \left( \boldsymbol{\omega} \right) 
\end{equation}

Using the log-derivative trick \cite{hoffman2013stochastic}, the objective function $\mathcal{L} \left( \boldsymbol{\omega}, \boldsymbol{\xi} \right)$ can be differentiated with respect to variational parameters $\boldsymbol{\xi}$:

\begin{equation}
\label{eq: sec3_19}
\begin{aligned}
\nabla_{\boldsymbol{\xi}} \mathcal{L} \left( \boldsymbol{\omega}, \boldsymbol{\xi} \right) & = \frac{\partial}{\partial \boldsymbol{\xi}} \int q_{\boldsymbol{\xi}} \left( \boldsymbol{\omega} \right) \mathcal{A} \left( \boldsymbol{\omega}, \boldsymbol{\xi} \right) d \boldsymbol{\omega} \\
& = \int q_{\boldsymbol{\xi}} \left( \boldsymbol{\omega} \right) \frac{\partial \log q_{\boldsymbol{\xi}} \left( \boldsymbol{\omega} \right)}{\partial \boldsymbol{\xi}} \mathcal{A} \left( \boldsymbol{\omega}, \boldsymbol{\xi} \right) + q_{\boldsymbol{\xi}} \left( \boldsymbol{\omega} \right) \frac{\partial \mathcal{A} \left( \boldsymbol{\omega}, \boldsymbol{\xi} \right)}{\partial \boldsymbol{\xi}} d \boldsymbol{\omega}
\end{aligned}
\end{equation}

To quickly estimate numerical integrations, we can write \cref{eq: sec3_19} in its expectatio form and use Monte Carlo method to compute the stochastic gradients:

\begin{equation}
\label{eq: sec3_20}
\nabla_{\boldsymbol{\xi}} \mathcal{L} \left( \boldsymbol{\omega}, \boldsymbol{\xi} \right) = \mathbb{E}_{q_{\boldsymbol{\xi}} \left( \boldsymbol{\omega} \right)} [ \frac{\partial \log q_{\boldsymbol{\xi}} \left( \boldsymbol{\omega} \right)}{\partial \boldsymbol{\xi}} \mathcal{A} \left( \boldsymbol{\omega}, \boldsymbol{\xi} \right) + \frac{\partial \mathcal{A} \left( \boldsymbol{\omega}, \boldsymbol{\xi} \right)}{\partial \boldsymbol{\xi}}]
\end{equation}

However, it is observed that crude MC estimator for $\nabla_{\boldsymbol{\xi}} \mathcal{L} \left( \boldsymbol{\omega}, \boldsymbol{\xi} \right)$ usually induces large variance \cite{kingma2013auto, rezende2014stochastic}. For this reason, stochastic gradient variational Bayes (SGVB) method is embraced to reduce the estimations' variance \cite{kingma2013auto}. Simply put, SGVB introduces an auxiliary variable $\boldsymbol{\epsilon}$ to the proxy distribution:

\begin{equation}
\label{eq: sec3_21}
q_{\boldsymbol{\xi}} \left( \boldsymbol{\omega} \right) = \int q_{\boldsymbol{\xi}} \left( \boldsymbol{\omega}, \boldsymbol{\epsilon} \right) d \boldsymbol{\epsilon} = \int q_{\boldsymbol{\xi}} \left( \boldsymbol{\omega} | \boldsymbol{\epsilon} \right) p \left( \boldsymbol{\epsilon} \right) d \boldsymbol{\epsilon}
\end{equation}

where conditional probability density function $q_{\boldsymbol{\xi}} \left( \boldsymbol{\omega} | \boldsymbol{\epsilon} \right)$ is formally defined as a Dirac delta function:

\begin{equation}
\label{eq: sec3_22}
q_{\boldsymbol{\xi}} \left( \boldsymbol{\omega} | \boldsymbol{\epsilon} \right) = \delta \left( \boldsymbol{\omega} - g \left( \boldsymbol{\xi}, \boldsymbol{\epsilon} \right) \right)
\end{equation}

and $g \left( \boldsymbol{\xi}, \boldsymbol{\epsilon} \right)$ is a differentiable transformation function that connects $\boldsymbol{\omega}$ and $\boldsymbol{\epsilon}$:

\begin{equation}
\label{eq: sec3_23}
\boldsymbol{\omega} = g \left( \boldsymbol{\xi}, \boldsymbol{\epsilon} \right)
\end{equation}

For instance, a simple choice for  $p \left( \boldsymbol{\epsilon} \right)$ is isotropic Gaussian distribution $\boldsymbol{\epsilon} \overset{i.i.d.}{\sim} p \left( \boldsymbol{\epsilon} \right) = \mathcal{N} \, \left( \boldsymbol{0}, \, \boldsymbol{I} \right)$, and the reparameterization can be achieved though $\boldsymbol{\omega} = \boldsymbol{\mu}_{\boldsymbol{\omega}} + \boldsymbol{\sigma}_{\boldsymbol{\omega}} \odot \boldsymbol{\epsilon}$. Therefore, substituting \cref{eq: sec3_21} and \cref{eq: sec3_22} into \cref{eq: sec3_20}, the pathwise estimator can be expressed as \cite{kingma2013auto}:

\begin{equation}
\label{eq: sec3_24}
\nabla_{\boldsymbol{\xi}} \mathcal{L} \left( \boldsymbol{\omega}, \boldsymbol{\xi} \right) = \mathbb{E}_{p \left( \boldsymbol{\epsilon} \right)} [ \frac{\partial \log p \left( \boldsymbol{\epsilon} \right) \mathcal{A} \left( g \left( \boldsymbol{\xi}, \boldsymbol{\epsilon} \right), \boldsymbol{\xi} \right)}{\partial \boldsymbol{\xi}} + \frac{\partial \mathcal{A} \left( g \left( \boldsymbol{\xi}, \boldsymbol{\epsilon} \right), \boldsymbol{\xi} \right)}{\partial \boldsymbol{\xi}}]
\end{equation}

Combining \cref{eq: sec3_18} and \cref{eq: sec3_24}, the final Monte Carlo estimator for the gradients can be written as:

\begin{equation}
\label{eq: sec3_25}
\nabla_{\boldsymbol{\xi}} \mathcal{L} \left( \boldsymbol{\omega}, \boldsymbol{\xi} \right) = \mathbb{E}_{p \left( \boldsymbol{\epsilon} \right)} [ \frac{\partial}{\partial \boldsymbol{\omega}} [ \log p \left( \boldsymbol{\omega} , \boldsymbol{\mathcal{D}} \right) - \log q_{\boldsymbol{\xi}} \left( \boldsymbol{\omega} \right) ] \frac{\partial g \left( \boldsymbol{\xi}, \boldsymbol{\epsilon} \right)}{\partial \boldsymbol{\xi}} ]
\end{equation}

Now, the variance of stochastic gradients can be effectively reduced by magnitude of orders using this reparameterized estimator \cite{kingma2013auto}, and the VI-based optimization problem can be efficiently solved by the stochastic gradient descent algorithm mentioned in the previous section.

\subsection{Predictive evaluation: Monte Carlo sampling}
\label{sec33}
The last but the most important step of Bayesian computation concerns making predictions for new data samples $(\boldsymbol{x^*}, \boldsymbol{y^*})$, where the predictive distribution can be expressed as:

\begin{equation}
\label{eq: sec3_26}
p \left( \boldsymbol{y^*} | \boldsymbol{x^*}, \boldsymbol{\mathcal{D}} \right) = \int p \left( \boldsymbol{y^*} | \boldsymbol{x^*}, \boldsymbol{\omega} \right) p \left( \boldsymbol{\omega} | \boldsymbol{\mathcal{D}} \right) d \boldsymbol{\omega}
\end{equation}

The optimized proxy posterior $q_{\boldsymbol{\xi}} \left( \boldsymbol{\omega} \right)$, which is obtained by solving the ELBO optimization problem, will take the place of the true posterior distribution $p \left( \boldsymbol{\omega} | \boldsymbol{\mathcal{D}} \right)$:

\begin{equation}
\label{eq: sec3_27}
p \left( \boldsymbol{y^*} | \boldsymbol{x^*}, \boldsymbol{\mathcal{D}} \right) \simeq \int p \left( \boldsymbol{y^*} | \boldsymbol{x^*}, \boldsymbol{\omega} \right) q_{\boldsymbol{\xi}} \left( \boldsymbol{\omega} \right) d \boldsymbol{\omega}
\end{equation}

In the same vein, the predictive integral is numerically achieved by drawing random samples from the proxy distribution. An unbiased estimator is given:

\begin{equation}
\label{eq: sec3_28}
p \left( \boldsymbol{y^*} | \boldsymbol{x^*}, \boldsymbol{\mathcal{D}} \right) \approx \frac{1}{k} \sum^{k}_{i=1} p \left( \boldsymbol{y^*} | \boldsymbol{x^*}, \boldsymbol{\omega}_{i} \right) \quad \text{where} \quad \boldsymbol{\omega}_{i} \sim q_{\boldsymbol{\xi}} \left( \boldsymbol{\omega} \right)
\end{equation}

For the purpose of uncertainty representation, it is of great importance to compute statistical moments of $\boldsymbol{y^*}$, such as mean:

\begin{equation}
\label{eq: sec3_29}
\hat{\boldsymbol{y}}^{*}_{mean} =  \frac{1}{k} \sum^{k}_{i=1} \hat{\mathcal{F}} \left( \boldsymbol{x^*} \, | \, \boldsymbol{\omega}^{i} \right)
\end{equation}

and variance:

\begin{equation}
\label{eq: sec3_30}
\begin{aligned}
\hat{\boldsymbol{y}}^{*}_{var} & = \frac{1}{k} \sum^{k}_{i=1} \left( \tau_{\boldsymbol{\epsilon}}^{i} \boldsymbol{I} + \left( \hat{\mathcal{F}} \left( \boldsymbol{x^*} \, | \, \boldsymbol{\omega}^{i} \right) \hat{\mathcal{F}}^{T} \left( \boldsymbol{x^*} \, | \, \boldsymbol{\omega}^{i} \right) \right) \right) \\
& - \left( \frac{1}{k} \sum^{k}_{i=1} \hat{\mathcal{F}} \left( \boldsymbol{x^*} \, | \, \boldsymbol{\omega}^{i} \right) \right) \left( \frac{1}{k} \sum^{k}_{i=1} \hat{\mathcal{F}} \left( \boldsymbol{x^*} \, | \, \boldsymbol{\omega}^{i} \right) \right)^{T}
\end{aligned}
\end{equation}

Because $\hat{\boldsymbol{y}}^{*}_{mean}$ and $\hat{\boldsymbol{y}}^{*}_{var}$ are essential elements for constructing the acquisition function which balances the exploration and exploitation in the context of Bayesian optimization \cite{rasmussen2004gaussian, snoek2015scalable}.

\section{Numerical examples and results}
\label{sec4}
\subsection{Example 1: nonlinear regression}
\label{sec41}
The first example considers a nonlinear function, which is commonly used as a testing problem to assess the accuracy of a regression model \cite{snoek2015scalable, goodfellow2016deep, springenberg2016bayesian}. Mathematically, it is written as:

\begin{equation}
\begin{aligned}
\label{eq: sec4_1}
y = x \, sin \left( x \right) 
\end{aligned}
\end{equation}
 
To identify common features and differences between proposed model and current approaches, the regression problem is numerically solved using four different surrogate modeling methods: polynomial regression (PR); support vector machine (SVM); neural networks (NN); and Bayesian deep learning (BDL). First, a polynomial $\hat{f} \left( x \right) = \beta_0 + \beta_1 x + \dots + \beta_n x^n$ of degree $n = 11$ is defined to fit the symmetric function \cite{nasrabadi2007pattern}. The method of least squares is applied to find the best linear unbiased estimator (BLUE) of the regression coefficient vector $\boldsymbol{\beta}$ by minimizing the sum of squared errors. Secondly, an SVM regression model with a Gaussian kernel function $G \left( x_i, x_j \right) = \exp \left( - \gamma || x_i - x_j ||^2 \right)$ is implemented to build a mapping between $x$ and $y$ \cite{robert2014machine, rasmussen2004gaussian}. The default value for the kernel coefficient $\gamma$ is $1$, and sequential minimal optimization (SMO) algorithm is utilized to update the coefficients where Karush-Kuhn-Tucker (KKT) violation $\epsilon = 0.0001$ is specified as the convergence criterion \cite{platt1998sequential}. Thirdly, a feedforward neural network with one hidden layer that has $20$ neurons is built to learn the nonlinear transformation \cite{goodfellow2016deep, nasrabadi2007pattern}. Hyperbolic tangent function is adopted as the activation function for the hidden layer since its derivatives are steeper than sigmoid function. For the output layer, a straight line function that outputs the weighted sum from hidden neurons is used. Stochastic gradient descent is performed for parameter optimization \cite{robbins1985stochastic}, where the learning rate is fixed as a constant $\eta = 0.0001$ and the default epoch setting is $100000$. Lastly, a Bayesian surrogate that has the same network configuration is examined. To account for the model uncertainty, a normal prior $\mathcal{N} \left( 0, 0.1 \right)$ is directly imposed on the model parameters.

\begin{figure}[ht]
\centering
\includegraphics[width=1.0\textwidth]{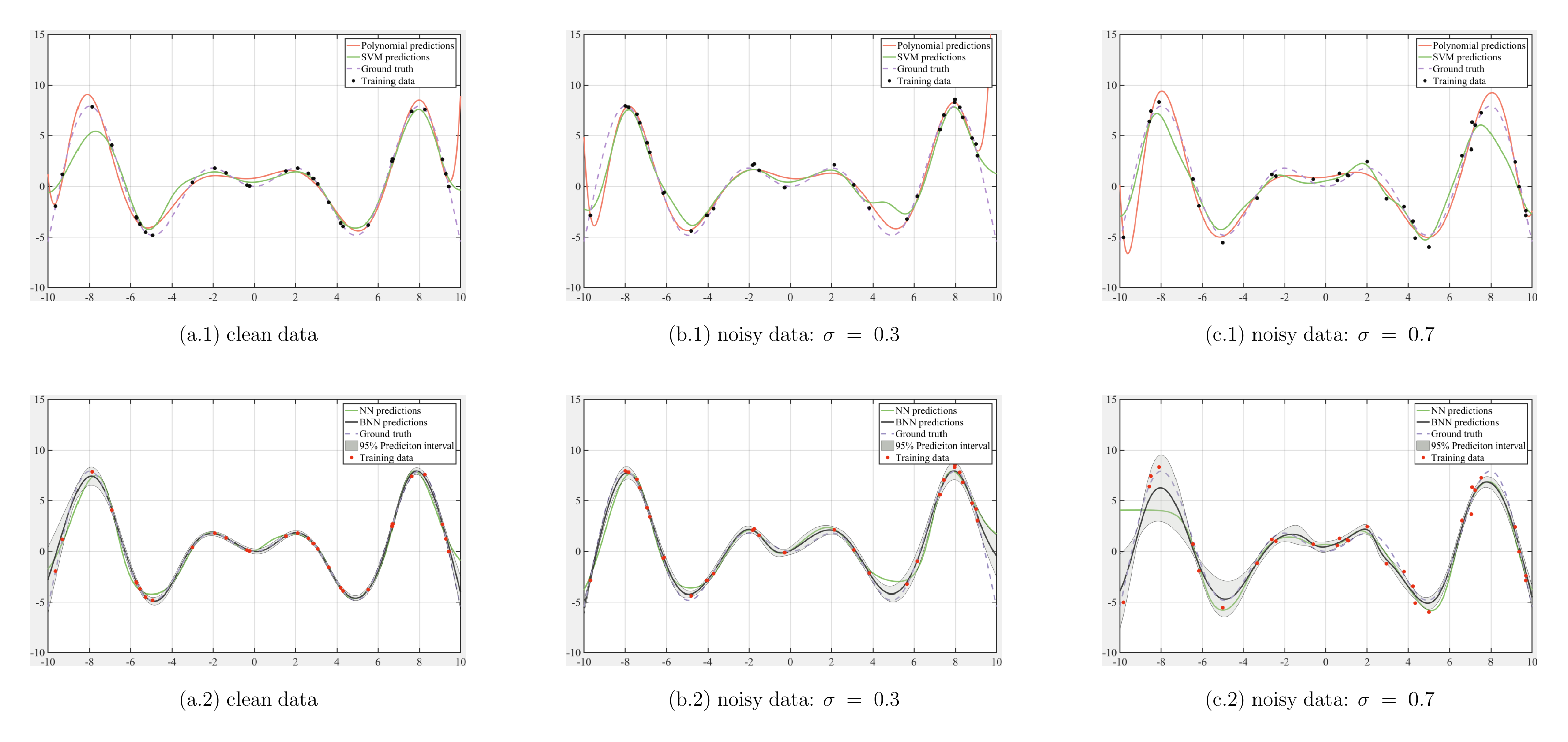}
\caption{Comparisons of regression results using various surrogate models. The training dataset is contaminated by a Gaussian noise with different standard deviations.}
\label{fig: f3}
\end{figure}

To train these models, we use the pseudorandom number generator to simulate a training dataset consisting of $30$ samples that are uniformly distributed in the interval $\left( -10, 10 \right)$. A Gaussian noise determined by $\epsilon_i \sim \mathcal{N} \left( 0, \sigma \right)$ is added to each sample to make the problem more realistic \cite{kendall2017uncertainties}. \cref{fig: f3} visualizes the fitted regression model via different approaches. It should be noted that the mean value of the predictive distribution is selected as the model estimation in the case of BDL. Obviously, BDL improves the generalization performance and mitigates the overfitting issue, which is encountered in NN modeling. Meanwhile, BDL is capable of characterizing the model uncertainty associated with the prediction in addition to achieving an equivalently accurate regression result compared to other methods. \cref{table: t1} and \cref{table: t2} summarize the coefficient of determination ($R^2$) and the root mean squared error (RMSE) for different surrogates. According to the results, BDL is more resistant to noisy data since increasing the random noise level deteriorates the effectiveness and quality of other three surrogates in a much more clear way.
 
\begin{table}[h]
\centering
\begin{tabular}{l l l l l l l}
\hline
\textbf{Method} & \textbf{clean data} & \textbf{$\boldsymbol{\sigma = 0.1}$} & \textbf{$\boldsymbol{\sigma = 0.3}$} & \textbf{$\boldsymbol{\sigma = 0.5}$} & \textbf{$\boldsymbol{\sigma = 0.7}$} & \textbf{$\boldsymbol{\sigma = 0.9}$}\\
\hline
PR & 0.9950 & 0.9937  & 0.9884  & 0.9807  & 0.9680  & 0.9533\\
SVM & 0.9783 & 0.9784  & 0.9758  & 0.9604  & 0.9486  & 0.9388\\
NN & 0.9890 & 0.9854  & 0.9828  & 0.9770  & 0.9526  & 0.9497\\
BDL & 0.9883 & 0.9893  & 0.9928  & 0.9757  & 0.9672  & 0.9516\\
\hline
\end{tabular}
\caption{Comparison of the coefficient of determination ($R^2$) of the different surrogate models where the training dataset is contaminated by different noise levels.}
 \label{table: t1}
\end{table}

\begin{table}[h]
\centering
\begin{tabular}{l l l l l l l}
\hline
\textbf{Method} & \textbf{clean data} & \textbf{$\boldsymbol{\sigma = 0.1}$} & \textbf{$\boldsymbol{\sigma = 0.3}$} & \textbf{$\boldsymbol{\sigma = 0.5}$} & \textbf{$\boldsymbol{\sigma = 0.7}$} & \textbf{$\boldsymbol{\sigma = 0.9}$}\\
\hline
PR & 0.2483 & 0.2629  & 0.3917  & 0.4863  & 0.6227  & 0.7643\\
SVM & 0.5397 & 0.5424  & 0.5642  & 0.6997  & 0.7934  & 0.9198\\
NN & 0.3817 & 0.4068  & 0.4783  & 0.5276  & 0.7350  & 0.8077\\
BDL & 0.3270 & 0.3098  & 0.2964  & 0.5425  & 0.6091  & 0.7933\\
\hline
\end{tabular}
\caption{Comparison of the root mean squared error (RMSE) of the different surrogate models where the training dataset is contaminated by different noise levels.}
\label{table: t2}
\end{table}

\subsection{Example 2: binary classification}
\label{sec42}
\begin{figure}[ht]
\centering
\includegraphics[width=0.9\textwidth]{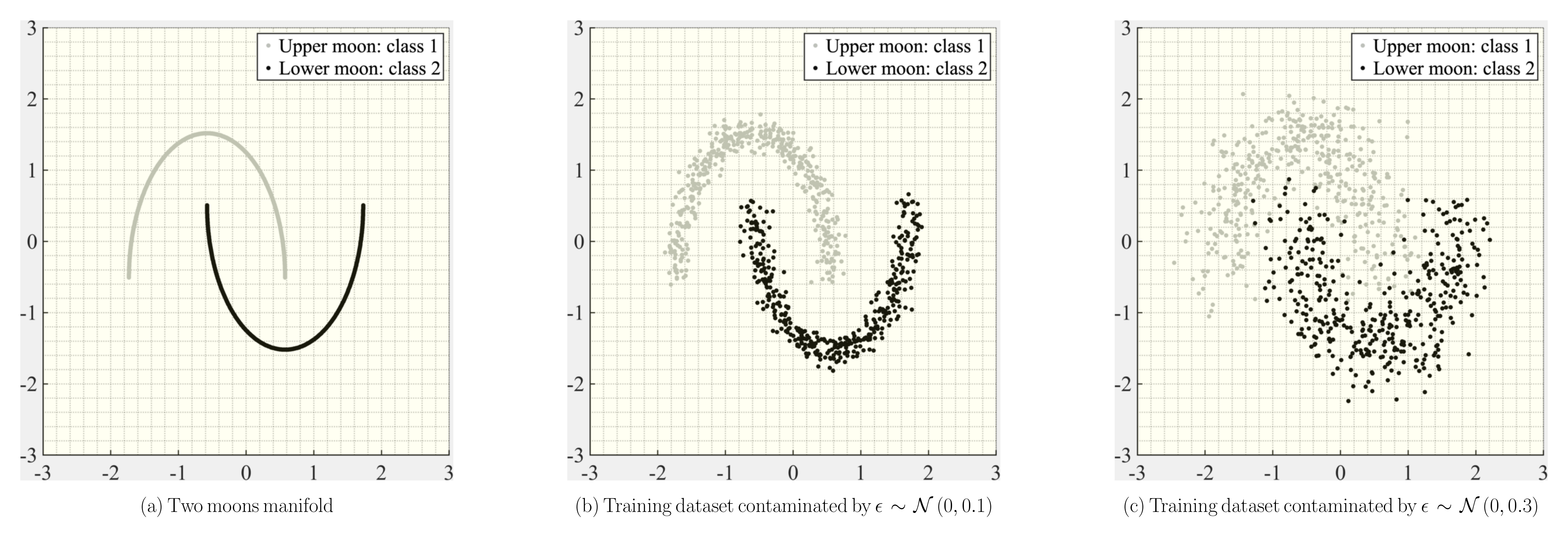}
\caption{Classification problem: a highly nonlinear dataset.}
\label{fig: f4}
\end{figure}

To evaluate the classification performance of our proposed surrogate model, the second example applies the BDL to a synthetic dataset that holds a two-dimensional swirl pattern. As shown in \cref{fig: f4}, the synthetic dataset exhibits two intuitively separable manifolds, where each manifold resembles a crescent moon \cite{robert2014machine}.

A BDL surrogate that is arranged in a $2 \times 5 \times 5 \times 2$ form is developed as the neural network classifier. Specifically, two hidden layers are configured with the hyperbolic tangent activation function and softmax function $\sigma \left( \boldsymbol{x}_i \right) = \frac{ e^{ \boldsymbol{x}_i }}{\sum_{j=1}^J e^{\boldsymbol{x}_j}}$ is implemented to represent the categorical distribution for the outputs by computing a probability row vector where the sum of the row is $1$ \cite{goodfellow2016deep, lecun2015deep}. To understand the effects of different priors on the classification performance, we have considered three direct priors: Laplace $\mathcal{L} \left( 0, 1 \right)$, Gaussian $\mathcal{N} \left( 0, 1 \right)$, and Cauchy $\mathcal{C} \left( 1, 1 \right)$. We further come up with three more hyper priors by fixing the location parameter $\alpha$ of the aforementioned probability distributions along with treating their scale parameter $\beta$ as a random variable, which can be described using an Inverse-Gamma distribution $\mathcal{IG} \left( 1, 1 \right)$. The basic probability distribution functions are given as:

\begin{equation}
\begin{aligned}
\label{eq: sec4_2}
\mathcal{L} \left( x \mid \alpha, \beta \right) & = \frac{1}{2 \beta} \exp \left\{ - \frac{|x - \alpha|}{\beta} \right\} \\
\mathcal{C}  \left( x \mid \alpha, \beta \right) & = \frac{1}{\pi \beta [1 + (\frac{x-\alpha}{\beta})^2]} \\
\mathcal{IG} \left( x \mid \alpha, \beta \right) & = \frac{\beta^{\alpha}}{\Gamma(\alpha)} x^{-\alpha - 1} \exp\left(\frac{-\beta}{x}\right)
\end{aligned}
\end{equation}

Additionally, we conduct two trials to study the effects of noise on our neural network classifier. In the first trial, a BDL model is developed using $900$ samples, where each sample is contaminated by a Gaussian noise generated from $\epsilon_i \sim \mathcal{N} \left( 0, 0.1 \right)$. In the second trial, $1200$ samples are utilized to build $\mathcal{M}^{'} \left( \cdot \right)$ as the external noise is amplified to $\epsilon_i \sim \mathcal{N} \left( 0, 0.3 \right)$. Following the $70/30$ rule \cite{goodfellow2016deep, lecun2015deep, nasrabadi2007pattern}, the whole dataset $\mathcal{D}$ is divided into the training set $\mathcal{D}_{t}$ and the validation set $\mathcal{D}_{v}$, respectively. A first-order gradient-based optimization method, ADAM \cite{kingma2014adam}, is adopted to update model parameters, where the learning rate $\eta = 0.001$, the exponential decay rates for the first/second moment estimates $\beta_1$ and $\beta_2$ are $0.9$ and $0.999$, respectively. It should be addressed that the reparametrization trick mentioned in \cref{sec32} is automatically embedded by means of taking the derivatives of the objective function with respect to the variational parameters \cite{kingma2013auto}. Here, the proxy distribution takes a Gaussian form, which indicates the variational posterior distribution is parameterized with two parameters, mean and standard deviation. To accelerate the training process, Mini-batch optimization method is used \cite{goodfellow2016deep, hoffman2013stochastic}, and the batch size is set to $30$. The stop criteria epoch number is $50000$. 

\begin{figure}[ht]
\centering
\includegraphics[width=1.0\textwidth]{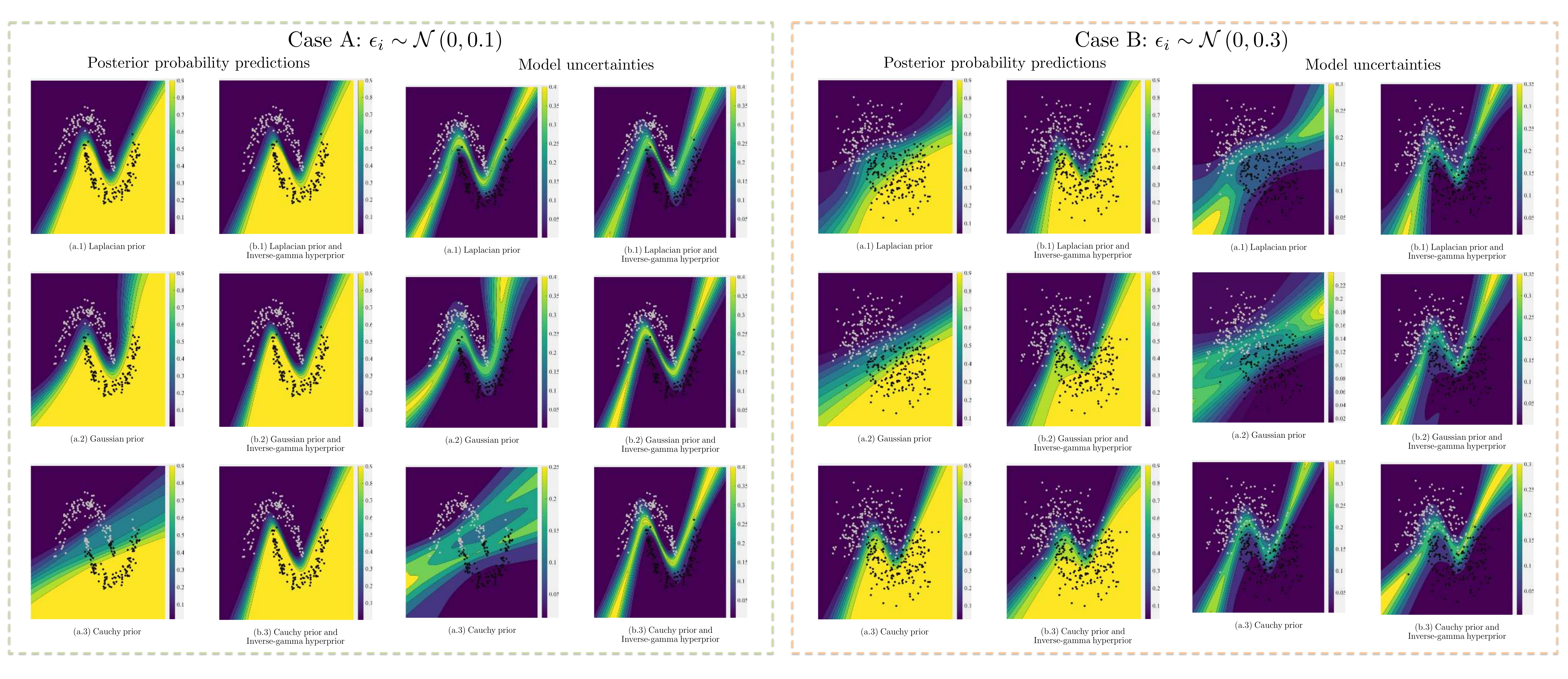}
\caption{Classification results: the predicted results are represented by the mean predictive probability of the lower crescent on the input domain of $\left( -3, 3 \right) \times \left( -3, 3 \right)$ and the model uncertainty is quantified in terms of the variance associated with each prediction using \cref{eq: sec3_30}.}
\label{fig: f5}
\end{figure}

\begin{figure}[ht]
\centering
\includegraphics[width=0.9\textwidth]{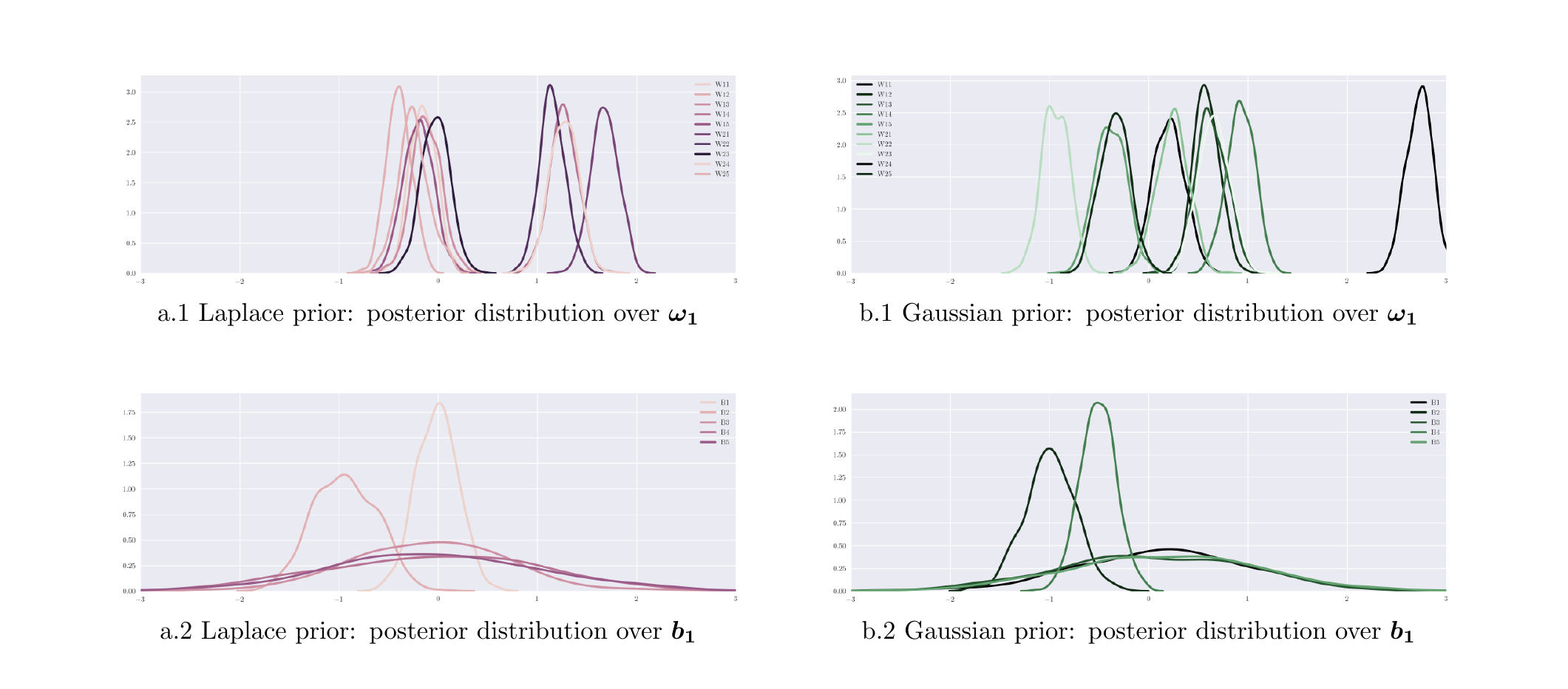}
\caption{Comparison of optimized variational posterior distributions of model parameters using different regularization techniques. $\boldsymbol{\omega_1}$ and $\boldsymbol{b_1}$ denote the weight and bias tensor associated with the first layer, respectively.}
\label{fig: f6}
\end{figure}

\cref{fig: f5} provides a graphic illustration of the classification results. According to these results, BDL model becomes less confident about its predictions when validation samples are more near the true separation trajectory. It is because even small noise can distort the original manifold in a severe way \cite{natarajan2013learning}. However, the proposed hyper priors are able to provide better predictions especially in the second trial where the addictive noise is stronger. This is credited to the nature mechanism of Bayesian hierarchical modeling, which relaxes the prior constraints by encoding prior belief using a series of hyperparameter values instead of fixed constants \cite{robert2014machine}. Lastly, \cref{fig: f6} reveals the variational posterior distribution of weights and bias in the first hidden layer. For the previous proposed priors, zero centered Laplace prior is equivalent to the $L1$ regularization and Gaussian prior is identical to the $L2$ regularization \cite{nalisnick2018priors, blundell2015weight}. In \cref{fig: f6}, results of $\mathcal{L} \left( 0, 1 \right)$ is approximately sparse signal and $p \left( \boldsymbol{\omega} \right)$ of $\mathcal{N} \left( 0, 1 \right)$ is not centering around zero, which aligns with the properties of $L1$ and $L2$ regularization respectively \cite{nasrabadi2007pattern}.

\subsection{Example 3: structural analysis of a geometrically nonlinear membrane}
\label{sec43}
\begin{figure}[ht]
\centering
\includegraphics[width=1.0\textwidth]{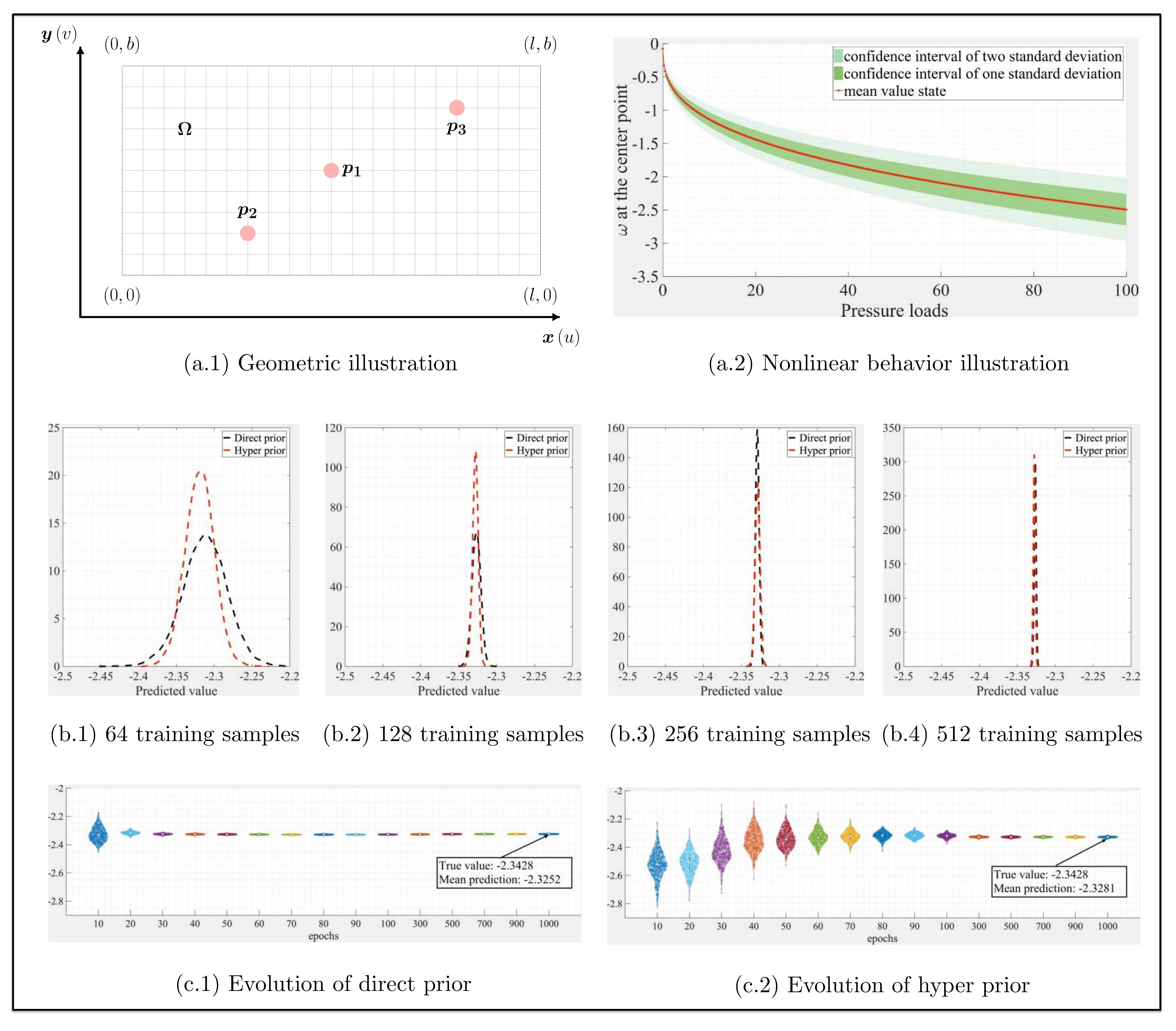}
\caption{Membrane example: problem statement and optimization results.}
\label{fig: f7}
\end{figure}

This example addresses the computational cost issue of using finite element (FE) model in the structural analysis with uncertain inputs \cite{Luo:inpress-a}. The target structure is a geometrically nonlinear membrane that is clamped at four edges \cite{kim2014introduction, kuether2014numerical}. \cref{fig: f7}. (a.1) gives a sketch of the objective domain $\Omega = [0, l] \times [0, b] \subset \mathbb{R}^2$, where uniformly distributed pressure loads are applied on the upper surface. We are interested in using the BDL based surrogate $\mathcal{M}^{'} \left( \cdot \right)$ to approximate the nonlinear mechanism between uncertain structure parameters $\boldsymbol{x}$ and random responses $\boldsymbol{y}$.

\subsubsection{Membrane example: nonlinear analysis and surrogate modeling}
\label{sec431}
\textbf{\textit{Uncertainty analysis.}} Vector $\boldsymbol{x}$ covers geometric uncertainties, where $x_1 = l$, $x_2 = b$, and $x_3 = t$ are the length, breadth, and thickness of target membrane, as well as material uncertainties, with $x_4 = E$ and $x_5 = \upsilon$ denoting the elastic modulus and Poisson's ratio, respectively. \cref{table: t3} gives a systematic summary of statistical properties of $\boldsymbol{x}$. The quantities of interest $\boldsymbol{y}$ are z-direction displacements $w$ at locations of $p_1 - \left(1, 0.5 \right)$, $p_2 - \left( 0.6, 0.2 \right)$, and $p_3 - \left(1.6, 0.8 \right)$. The load-displacement relationship is no longer a deterministic curve due to the input randomness (See \cref{fig: f7}. (a.2)).

\begin{table}[ht]
\begin{threeparttable}
\begin{tabular}{l l l l}
\hline
\textbf{Basic variables} & \textbf{First parameter} & \textbf{Second parameter} & \textbf{Distribution type}\\
\hline
$l$ & $\mu = 2$ & $\sigma = 0.05$ & Normal \\
$b$ & $\mu = 1$ & $\sigma = 0.05$ & Normal \\
$t$ & $min = 0.001$ & $max = 0.002$ & Uniform \\
$E$ & $\mu = 210$ & $\sigma = 10$ & Normal \\
$\upsilon$ & $\mu = \log (0.3)$ & $\sigma = 0.01$ & Lognormal \\
\hline
\end{tabular}
\end{threeparttable}
\caption{Statistics of the uncertain input parameters for the flat membrane.}
\label{table: t3}
\end{table}

\textbf{\textit{Nonlinear FE model.}} Because of the geometric nonlinearity, the inplain strain is partitioned into two parts $\boldsymbol{\epsilon} = \boldsymbol{\epsilon}^l + \boldsymbol{\epsilon}^{non}$, where $\boldsymbol{\epsilon}^l$ describes the linear strain and $\boldsymbol{\epsilon}^{non}$ represents the nonlinear strain term:

\begin{equation}
\label{eq: sec4_4}
\begin{aligned}
\boldsymbol{\epsilon}^{non} = \begin{bmatrix} \epsilon_x^{non} \\ \epsilon_y^{non} \\ \gamma_{xy}^{non} \end{bmatrix} = \frac{1}{2} \begin{bmatrix} \left( \frac{\partial{u}}{\partial{x}} \right)^2 + \left( \frac{\partial{v}}{\partial{x}} \right)^2 + \left( \frac{\partial{w}}{\partial{x}} \right)^2 \\ \left( \frac{\partial{u}}{\partial{y}} \right)^2 + \left( \frac{\partial{v}}{\partial{y}} \right)^2 + \left( \frac{\partial{w}}{\partial{y}} \right)^2 \\ 2 \left( \frac{\partial{u}}{\partial{x}} \frac{\partial{u}}{\partial{y}} \right) + 2 \left( \frac{\partial{v}}{\partial{x}} \frac{\partial{v}}{\partial{y}} \right) + 2 \left( \frac{\partial{w}}{\partial{x}} \frac{\partial{w}}{\partial{y}} \right)\end{bmatrix}
\end{aligned}
\end{equation}

The solution $\boldsymbol{d} = [u, v, w]$ of these nonlinear equilibrium equations are obtained by the Newton-Raphson (NR) method \cite{kim2014introduction}. The iterative process terminates when the unbalanced force residual is smaller than the tolerance $\epsilon = 0.0001$ or the NR algorithm reaches the default maximum iteration $n = 100$. The force and displacement vector is initialized to zero and the increment loads $\Delta_{p} = p/n$ where $p = 100$ and $n=400$. It should be noted that the thickness of the membrane is comparatively small in relation to other two dimensions. It is therefore the FE model $\mathcal{M} \left( \cdot \right)$ can be built by 200 $(20 \, \text{in x axis} \times 10 \, \text{in y axis})$ four node (Q4) quadrilateral elements and large deformation theory is adopted \cite{kim2014introduction}.

\textbf{\textit{Surrogate model.}} We use the proposed BDL approach to provide a $\mathbb{R}^{5} \rightarrow \mathbb{R}^{3}$ transformation. The network architecture has three hidden layers $30 \times 15 \times 10$ besides the input and output layer \cite{abadi2016tensorflow} and probability distributions that account for model uncertainties are specified in a layer by layer fashion. To investigate the efficacy of different prior, a direct zero-mean Gaussian $\mathcal{N} \left( 0, 1 \right)$ and a hierarchical prior $\mathcal{N} \left( 0, \mathcal{IG} \left( 1, 1 \right) \right)$ have been tested. Furthermore, training datasets of size $64$, $128$, $256$, and $512$ have been considered for the purpose of identifying the influence from the amount of data on the accuracy of model predictions. For the posterior approximation, ADAM is adopted \cite{kingma2014adam}, where the initial learning rate $\eta$ is $0.005$. Notably, $\eta$ decays every $100$ epochs by multiplying a constant rate of $0.75$ and the epoch number is $1000$. The batch size for the subsampling procedure of all trials is set to $16$. In the variational inference stage, $200$ numerical samples are employed to estimate the lower bound.

\subsubsection{Results}
\label{sec432}
\begin{figure}[ht]
\centering
\includegraphics[width=0.9\textwidth]{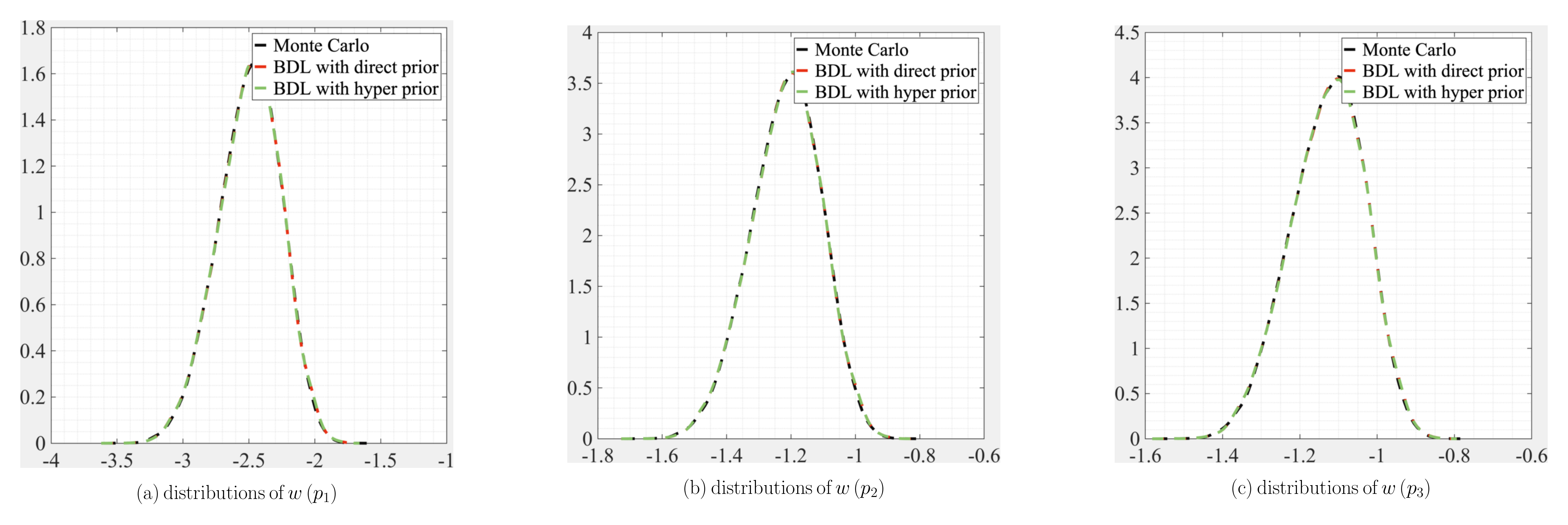}
\caption{Distribution estimate for $w \left( p_1 \right)$, $w \left( p_2 \right)$, and $w \left( p_3 \right)$. The dashed black line is the ground truth, which is computed via the high-fidelity FE model using $5 \times 10^5$ samples. The color lines denote the surrogate predictions that are kernel smoothing function estimates using the predictive mean.}
\label{fig: f8}
\end{figure}

In \cref{fig: f7}, the optimization results imply the predictive distribution computed by \cref{eq: sec3_28} shrinks rapidly as the number of training sample increases. $128$ samples can give a narrow-band distribution of $w \left( p_1 \right)$, indicating the trained BDL model becomes sufficiently reliable as most model uncertainties have been explained away by data. Moreover, we compared the evolution process of the predictive distribution $p \left( w_{p_1} \right)$ via different priors. In both trials, $128$ training samples are used, and the same validation sample is randomly chosen where $\boldsymbol{x} = [ 2.0117 ,   1.0157,    0.0019,  213.1180,    0.3018 ]$ and $\boldsymbol{y} = [-2.3428,   -1.1501,   -1.0536 ]$. It is found that the predictive distribution via hyper prior takes more epochs to shrink (See \cref{fig: f7}. (c.1) and (c.2)). The intuitive explanation is $\mathcal{N} \left( 0, \mathcal{IG} \left( 1, 1 \right) \right)$ has a larger initial parameter space than $\mathcal{N} \left( 0, 1 \right)$. Despite the difference, both BDL surrogates provide a reliable input-output mapping, and the coefficient of determination is summarized in \cref{table: t4}. To check the generalizability of proposed surrogates, $\mathcal{M}^{'} \left( \cdot \right)$ is further applied to uncertainty analysis. First, $1000$ samples were used to train the network. Next, another $1 \times 10^5$ samples are exploited to develop the distribution of displacements at $p_1$, $p_2$, and $p_3$. \cref{fig: f8} presents the UQ results, where BDL based surrogates accurately propagate uncertainties to the response distributions. 

\begin{table}[h]
\centering
\begin{tabular}{l l l l l}
\hline
\textbf{Prior type} & $\{ \boldsymbol{x}_i, \boldsymbol{y}_i \}_{i=1}^{64}$ & $\{\boldsymbol{x}_i, \boldsymbol{y}_i\}_{i=1}^{128}$ & $\{\boldsymbol{x}_i, \boldsymbol{y}_i\}_{i=1}^{256}$ & $\{\boldsymbol{x}_i, \boldsymbol{y}_i\}_{i=1}^{512}$\\
\hline
Direct prior & 0.9720 & 0.9982 & 0.9990 & 0.9993 \\
Hyper prior & 0.9935 & 0.9989 & 0.9983 & 0.9994 \\
\hline
\end{tabular}
\caption{Comparison of the coefficient of determination.}
\label{table: t4}
\end{table}

\subsection{Example 4: prediction of wind pressure}
\label{sec44}
Obtaining detailed data of wind-induced pressure coefficients on building surfaces is of great practical importance in the design of high-rise buildings. However, wind tunnel test results are limited and may be contaminated through different sources. In this example, the proposed BDL model is applied to predict the mean and root-mean-square (RMS) pressure coefficients using limited experiment data.

\subsubsection{Wind pressure database and predictive model}
\label{sec441}
\textbf{\textit{Wind tunnel data.}} The aerodynamic database considered for this example is developed by the Tokyo Polytechnic University (TPU) \cite{TPU}. For the wind tunnel experiment, a $1:400$ scale rigid model was built to represent the target tall building of dimension $200 m \times 40 m \times 40 m$ and a power law exponent of $1/4$ was used for the description of the mean wind speed. A total of $500$ pressure taps were used to collect data at a sampling frequency of $1000 \, Hz$ for a sample period of $32.768 \, s$. Hourly average wind speed was $11.1438 \, m/s$ and wind attacking angle was $0 ^{\circ}$, indicating wind direction is perpendicular to the front face of the experiment model. The wind pressure of our interest is characterized by a dimensionless number $\boldsymbol{C_p} \left( \boldsymbol{x} \right) = \frac{p_{\boldsymbol{x}} - p_{\infty}}{\rho U_0^2 /2}$ that is known as the pressure coefficient. $p_{\infty}$ is the static pressure at freestream, $\rho$ is the air density and $U_0$ is the mean wind speed at the reference height. The predictive quantities are:

\begin{equation}
\label{eq: sec4_6}
\begin{aligned}
\boldsymbol{C_p}^{mean} \left( \boldsymbol{x} \right) & = \mathbb{E} [ \boldsymbol{C_p} \left( \boldsymbol{x} \right) ] \\
\boldsymbol{C_p}^{rms} \left( \boldsymbol{x} \right) & = \sqrt{\frac{1}{N} | \boldsymbol{C_p} \left( \boldsymbol{x} \right) - \boldsymbol{C_p}^{mean} \left( \boldsymbol{x} \right) |^2}
\end{aligned}
\end{equation}

To demonstrate the efficacy of the BDL surrogate in dealing with small datasets, Biharmonic spline interpolation is performed based on the measured pressure data $N_{old} = 500$. Specifically, the are $250$ width interpolation points and $750$ height interpolation points in each building face. As a result, the surface pressure fields are described by a total of $N_{new} = 250 \times 750 \times 4 = 750000$ synchronous pressure points. However, we only use as much as $1 \%$ of the total data to train the Bayesian model.

\textbf{\textit{Prediction model.}} The Cartesian coordinates $\boldsymbol{x} \in \mathbb{R}^{2}$ are selected input variables of the BDL model, and the output is a scalar either $y = C_p^{mean} \left( \boldsymbol{x} \right)$ or $y = C_p^{rms} \left( \boldsymbol{x} \right)$. After extensive hyperparameters and network architectures search, it was found that BDL with network configuration of $2 \times 15 \times 10 \times 1$ and $2 \times 30 \times 15 \times 1$ provide superior performance in predicting $C_p^{mean}$ and $C_p^{rms}$, respectively. Hyperbolic tangent function $tanh \left( x \right) = \frac{e^{x} - e^{-x}}{e^{x} + e^{-x}}$ is adopted as the activation function since it produces steep derivatives, and ADMA optimizer \cite{kingma2014adam} is implemented with a learning rate initialized to $0.03$, which follows a step decay to prevent optimizing parameters chaotically. The annealing strategy is adopted to improve the stochastic gradients where the anneal rate is fixed to $0.75$. $1000$ samples are applied to get a reasonable estimation for the test log likelihood during the training phase. Epoch number is set as $1000$ and the testing frequency is $10$. To validate the effectiveness of HBM based prior, a direct Gaussian prior $\mathcal{N} \left( 0, 1 \right)$ and a hyper prior $\mathcal{N} \left( 0, \mathcal{IG} \left( 1, 1 \right) \right)$ are examined.

\subsubsection{Results}
\label{sec442}
\begin{figure}[ht]
\centering
\includegraphics[width=0.8\textwidth]{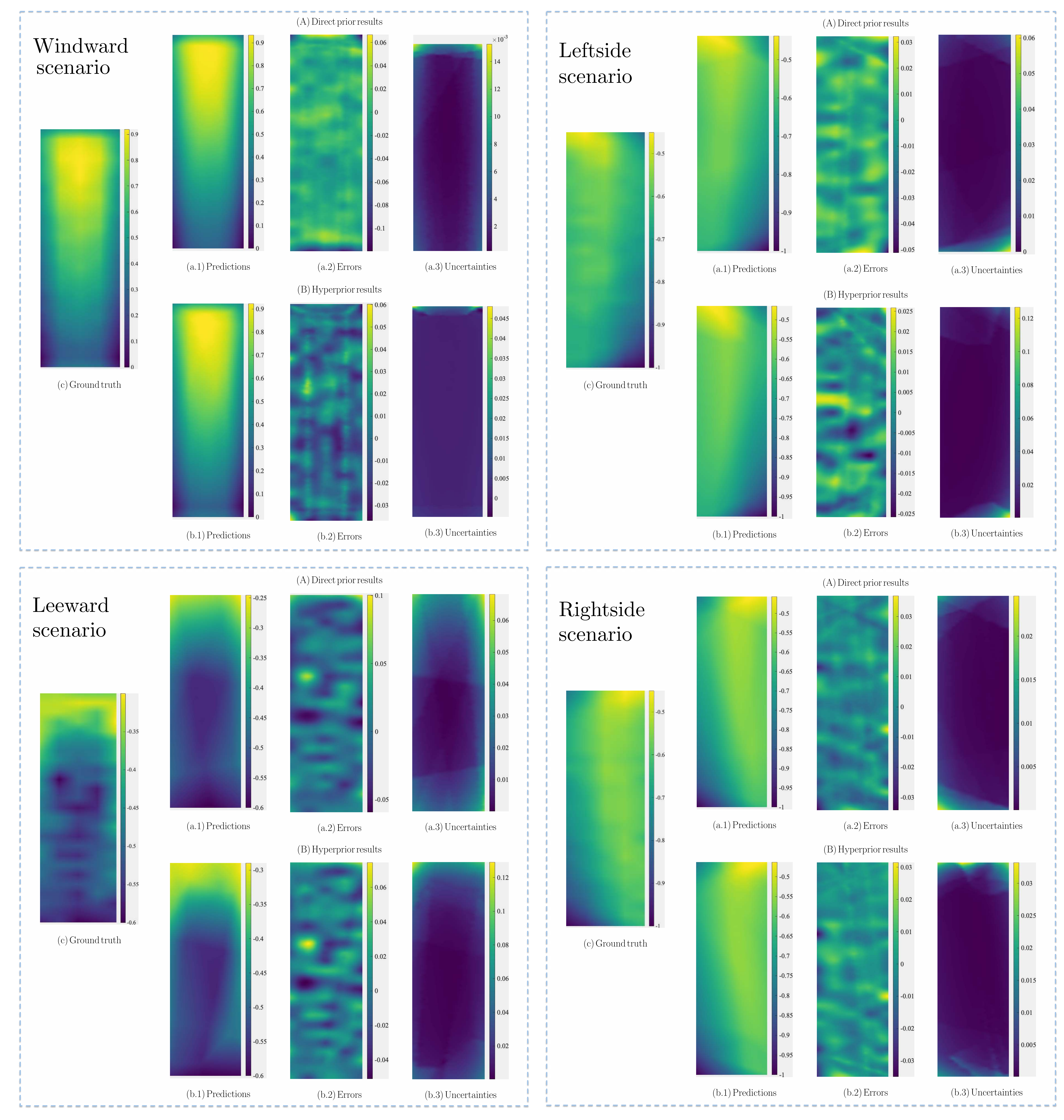}
\caption{Prediction results of $\boldsymbol{C_p}^{mean}$ using different priors. The training dataset has $700$ samples.}
\label{fig: f9}
\end{figure}

\begin{figure}[ht]
\centering
\includegraphics[width=0.8\textwidth]{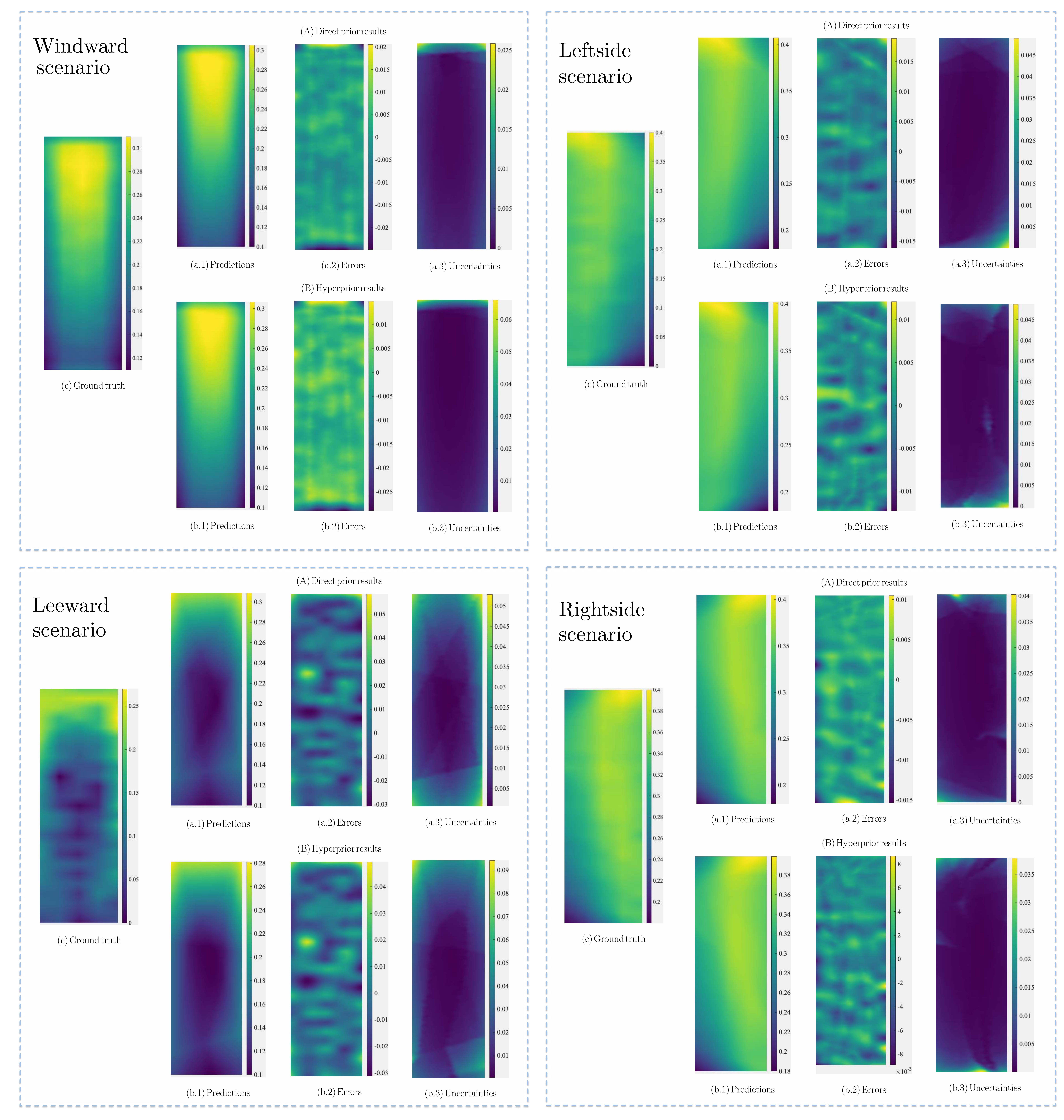}
\caption{Prediction results of $\boldsymbol{C_p}^{rms}$ using different priors. The training dataset has $700$ samples.}
\label{fig: f10}
\end{figure}

\begin{figure}[ht]
\centering
\includegraphics[width=0.9\textwidth]{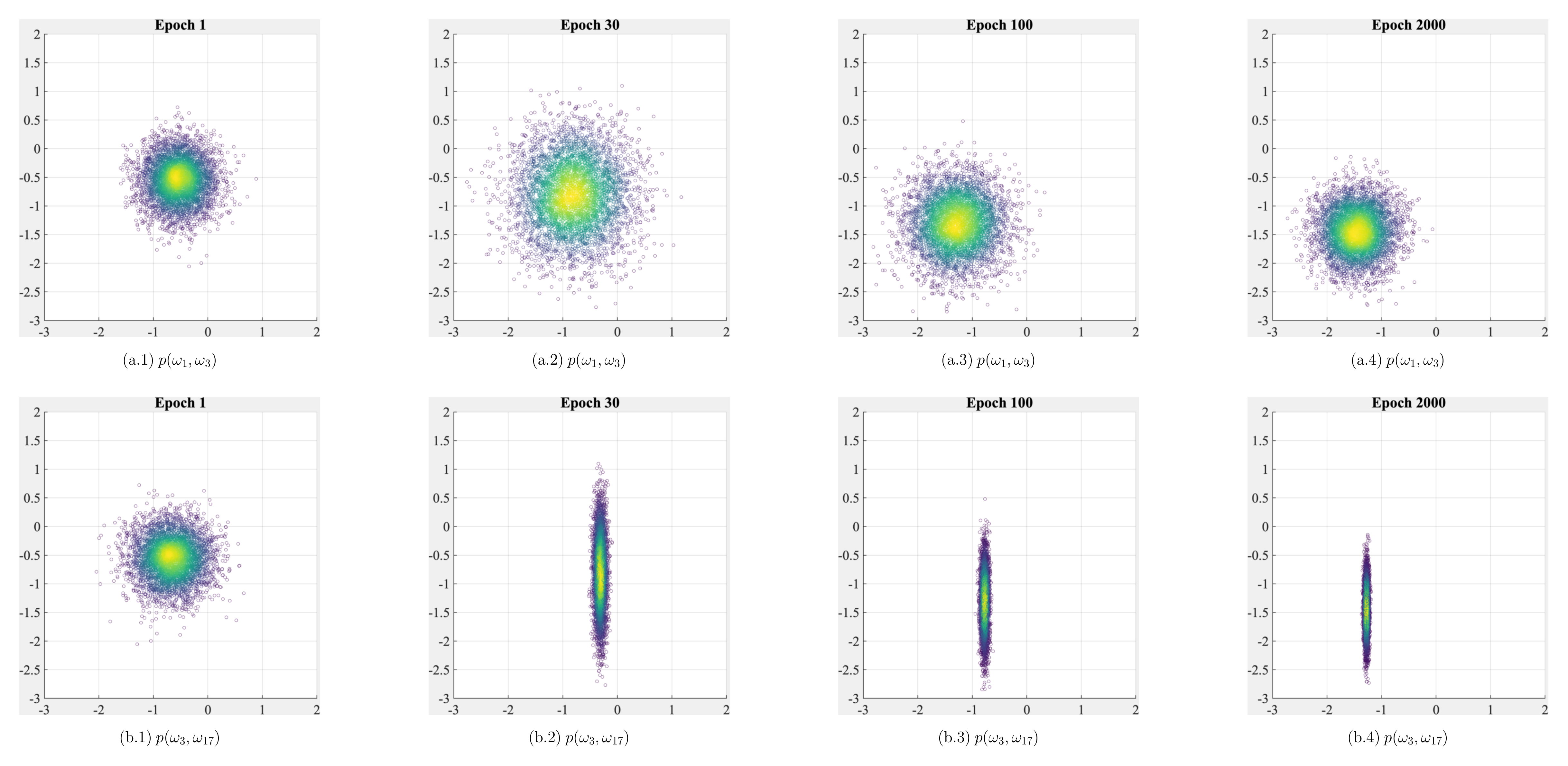}
\caption{Evolution of the variational posterior distribution. Three model parameters $\omega_1$, $\omega_3$, and $\omega_{17}$ are randomly selected. The first row corresponds to the joint distribution of $p \left( \omega_{1}, \omega_{3}\right)$, and the second row plots the joint distribution of $p \left( \omega_{3}, \omega_{17}\right)$. }
\label{fig: f11}
\end{figure}

\cref{fig: f9} summarizes the $\boldsymbol{C_p}^{mean}$ predictions and \cref{fig: f10} provides the predictive results of $\boldsymbol{C_p}^{rms}$. The ground truth is numerically obtained via the interpolation and extrapolation of the wind tunnel data $\boldsymbol{C_p}$; the predictions are defined as the mean value of the predictive distribution $\hat{\boldsymbol{C_p}}$; the relative error measures the predictive difference between $\boldsymbol{C_p}$ and $\hat{\boldsymbol{C_p}}$; and model uncertainties are evaluated using the variance of $\hat{\boldsymbol{C_p}}$. The results reveal that hyperprior $\mathcal{N} \left( 0, \mathcal{IG} \left( 1, 1 \right) \right)$ not only allows $\mathcal{M}^{'}\left( \cdot \right)$ to be better fitted but also makes $\mathcal{M}^{'}\left( \cdot \right)$ less sensitive to the complex nature of the noise embedded in the experimental data. It can be seen from the limits of the colorbar that the magnitude of relative errors is reduced for those hyperprior trials. From a percent error perspective, most predictions are lying within the $97 \%$ confidence interval and the maximum errors are all less than $10 \%$, most of which are mainly gathered around the boundary due to the extrapolation algorithm that is performed at the data preprocessing stage. In \cref{fig: f9}, uncertainties results confirm the propagation of uncertainty resulting from the extrapolation process and successfully identify the areas that are prone to. In general, it seems hyperprior improves the model performance more in the case of predicting $\boldsymbol{C_p}^{rms}$ than $\boldsymbol{C_p}^{mean}$, which is reasonable as the second sample moment magnifies the differences between predicted values and observed values, causing the need of more degrees of freedom to explain the variation. \cref{table: t5} summarizes the model performance using RMSE. To illustrate the robustness of the variational inference method in terms of approximating intractable posterior distributions, \cref{fig: f11} shows the optimization process of computing the variational posterior. It can be observed that VI is capable of capturing the main characteristics within the first $30$ epochs, and it is applicable to various situations where the true posterior distribution takes different kinds of forms.

\begin{table}[h]
\centering
\begin{tabular}{l l l l l l l}
\hline
\textbf{Scenario} & $\{ \boldsymbol{x}_i, \boldsymbol{y}_i \}_{i=1}^{100}$ & $\{\boldsymbol{x}_i, \boldsymbol{y}_i\}_{i=1}^{200}$ & $\{\boldsymbol{x}_i, \boldsymbol{y}_i\}_{i=1}^{300}$ & $\{\boldsymbol{x}_i, \boldsymbol{y}_i\}_{i=1}^{400}$ & $\{\boldsymbol{x}_i, \boldsymbol{y}_i\}_{i=1}^{500}$ & $\{\boldsymbol{x}_i, \boldsymbol{y}_i\}_{i=1}^{600}$\\
\hline
$\text{Windward}^{\dagger}$ & 0.077 & 0.030  & 0.027  & 0.022 & 0.019 & 0.017 \\
$\text{Windward}^{\ddagger}$ & 0.051 & 0.021  & 0.021  & 0.020 & 0.018 & 0.015 \\
$\text{Leftside}^{\dagger}$ & 0.039 & 0.030  & 0.027  & 0.019  & 0.013  & 0.012 \\
$\text{Leftside}^{\ddagger}$ & 0.037 & 0.030  & 0.024 & 0.016  & 0.012 & 0.012 \\
$\text{Leeward}^{\dagger}$ & 0.014 & 0.013  & 0.011  & 0.011  & 0.010  & 0.010 \\
$\text{Leeward}^{\ddagger}$ & 0.014 & 0.011  & 0.011  & 0.010  & 0.010  & 0.009 \\
$\text{Rightside}^{\dagger}$ & 0.037 & 0.025 & 0.025 & 0.021  & 0.019  & 0.015 \\
$\text{Rightside}^{\ddagger}$ & 0.033 & 0.025 & 0.023  & 0.020  & 0.014  & 0.011 \\
\hline
\end{tabular}
\begin{tablenotes}\footnotesize
\item ${\dagger}$ denotes the direct prior $\mathcal{N} \left( 0, 1 \right)$, and ${\ddagger}$ denotes the hyper prior $\mathcal{N} \left( 0, \mathcal{IG} \left( 1, 1 \right) \right)$
\end{tablenotes}
\caption{Comparison of the root mean square error.}
\label{table: t5}
\end{table}

\section{Concluding Remarks}
\label{sec5}
This paper presents a probabilistic modeling approach for learning hidden relationships from limited and noisy data using Bayesian deep learning (BDL) with hierarchical prior. The proposed surrogate rigorously accounts for the model uncertainties by means of imposing prior distributions on model parameters. Meanwhile, it effectively propagates the preassigned prior belief to the prediction quantities. In summary, the following conclusions are drawn:

\begin{itemize}
\item[(1)] Bayesian inference has been successfully integrated into the current deterministic deep learning framework. Consequently, the proposed model is able to analyze uncertainties associated with model predictions and help stakeholders make a more informed decision by providing a confidence level for the predictive estimation.
\item[(2)] The hypothesis of using hierarchical Bayesian modeling to describe prior distributions of model parameters is tested. In both classification and regression problems, superior performances can be achieved utilizing hyper priors, especially when the training data is seriously contaminated. Moreover, probabilistic surrogate with hyper prior tends to have an improved learning ability from a small dataset.
\item[(3)] Intractable posterior distributions that risen from multidimensional integrals step of Bayesian analysis has been addressed by the state-of-the-art variational inference method. Compared to some advanced sampling-based methods, variational inference method offers a higher scalability by tackling the model learning problem in an objective-equivalent-transformed and gradients-effective-computed optimization form.
\item[(4)] The examples provided have demonstrated the applicability of the proposed modeling scheme to both classification and regression tasks involving complex systems. Especially in the membrane example, BDL is capable of providing an accurate description of the highly nonlinear mapping between different design variables and various structural performance indicators and produces virtually identical uncertainty quantification results as conventional Monte Carlo method. Furthermore in the wind field prediction example, BDL model is trained and tested using very limited wind tunnel data, and it is shown that the probabilistic model is not only able to effectively recover the entire mapping of the mean and root-mean-square pressure fields with high precision using as small as $1\%$ of the data but also quantifies the uncertainty level at every single point in the prediction domain, serving as a reliable surrogate for learning complex field distribution.
\end{itemize}

To improve the model performance, it is envisaged that the combination of the BDL model with information from underlying physics can not only further accelerate the training of neural networks but also holds the promise of interpreting the learning process. 

\section{Acknowledgments}
\label{sec6}
This work was supported by the National Science Foundation (NSF) under Grant No. 1520817 and No. 1612843. This support is gratefully acknowledged.

\section{Reference}
\label{sec7}
\bibliographystyle{elsarticle-num}
\bibliography{manuscript}

\end{document}